\documentclass[12pt,onecolumn]{IEEEtran}
\usepackage{bm}
\usepackage{algorithm}
\usepackage{amssymb}
\usepackage{subcaption}
\usepackage{cite}
\usepackage{graphicx}
\usepackage{setspace}
\usepackage{color}
\usepackage[normalem]{ulem}
\usepackage{amsmath}
\usepackage{algorithmic}
\newcommand*\diff{\mathop{}\!\mathrm{d}}

\begin{document}
%
\title{{Reconstruction-Aware Imaging System Ranking by use of a Sparsity-Driven Numerical Observer Enabled by} Variational Bayesian Inference}
%
%
%

\author{Yujia Chen,~
        Yang Lou,~
        Kun Wang,~
        Matthew A. Kupinski,~
        and Mark A. Anastasio~
\thanks{This research was supported in part by NIH awards EB020168 and EB020604.}
\thanks{Yujia Chen, Yang Lou, Kun Wang and Mark A. Anastasio are with the Department
of Biomedical Engineering, Washington University in St Louis, St Louis,
MO, 63130 USA (e-mail: chen.yujia@wustl.edu, louy@wustl.edu, anastasio@wustl.edu).}
\thanks{Matthew A. Kupinski is with College of Optical Sciences, University of Arizona, Tucson, AZ, 85721 USA (e-mail:  mkupinski@optics.arizona.edu).}
}

\maketitle

\begin{abstract}
It is widely accepted that optimization of imaging system performance should be guided by task-based measures of image quality
 (IQ). 
{It has been advocated that imaging hardware or data-acquisition designs
 should be optimized by use of an ideal observer (IO) that exploits full statistical knowledge of the measurement
noise and class of objects to be imaged, without consideration of the reconstruction method.} In practice, accurate and tractable models of the complete object statistics are often difficult to determine. Moreover, in imaging systems that employ compressive sensing concepts, {imaging hardware and sparse image reconstruction are innately coupled technologies}.
In this work,  a {sparsity-driven observer (SDO)} that
 can be employed to optimize hardware by use of a stochastic object model describing object sparsity is described and investigated.
 The {SDO} and sparse reconstruction method can therefore be ``matched" in the sense that they both utilize the same statistical information regarding the class of objects to be imaged. To efficiently compute the {SDO} test statistic,
 computational tools developed recently for variational Bayesian inference with sparse linear models are adopted.
{The use of the SDO to rank data-acquisition designs in a stylized example as motivated by 
 magnetic resonance imaging (MRI) is demonstrated.
 This study reveals that the SDO can produce rankings that are consistent with visual
assessments of the reconstructed images but different from those produced by use of the traditionally employed Hotelling observer (HO).}

\end{abstract}


%
\IEEEpeerreviewmaketitle

\section{Introduction}
%
%
%
%
 The development of methodologies for optimizing  imaging system performance remains an
important and actively investigated topic. It is widely accepted that optimization of imaging system performance should be guided by task-based measures of image quality (IQ)
\cite{barrett2013foundations,metz1995toward,vennart1997icru,wagner1985unified,he2013model,gang2015task,sidky2008depth}.
 Signal detection tasks \cite{clarkson2000approximations,sanchez2014task}, such as tumor detection, are ubiquitous and relevant to a wide range of medical imaging modalities \cite{swets2014signal}. An upper performance bound for a binary detection task is achieved by the Bayesian ideal observer (IO) \cite{barrett2013foundations,clarkson2000approximations,geisler2003ideal}, which is a numerical observer that employs complete knowledge of the object and the noise statistics. Henceforth, any reference to IOs refers to Bayesian IOs.

It has been widely advocated to use IO performance, evaluated by use of an ensemble of
experimental or simulated  measurement data, as
a figure-of-merit (FOM) to guide optimization of imaging hardware and
 data-acquisition parameters for signal{-}detection tasks.
 A rationale for this is that the IO detection performance can be interpreted as a
measure of the ability of the imaging system to record a signal optimally distinguished from the background
structures and measurement noise.   
The IO has also been employed to assess the efficiency of human observers on signal
detection tasks\cite{liu1995object}.


{The IO performs the likelihood-ratio test for a classification task and thus the test statistic is a ratio of posterior probability distributions under competing hypotheses. This alone is a substantial impediment to using the true IO because it requires knowledge of the full probability density function (PDF) that describes the ensemble of objects to be imaged. Henceforth, this PDF will be referred to as a stochastic object model (SOM). While SOMs have been proposed for applications that include nuclear medicine imaging\cite{kupinski2003experimental,gross2005fast} and breast imaging\cite{graff2016new}, accurate and tractable models of object statistics generally are not available for most imaging applications. Moreover, even if a suitable SOM is available, computation of the likelihood ratio, in general, is not analytically tractable and remains computationally burdensome.}


 Markov chain
Monte Carlo (MCMC) techniques \cite{kupinski2003ideal} have been proposed to compute IO performance for a specific class of
object model such as the lumpy or clustered lumpy object models or parameterized phantoms\cite{he2008toward}. {In principle, the MCMC methods can provide unbiased results; however, their application is limited by long computation time, sometimes extending to weeks. 
Another disadvantage is that considerable expertise is required to run the MCMC simulations 
properly.} We are unaware of any convergence diagnostics that are simple to use or even widely
accepted. In practice, due to these complications, application of MCMC methods
 has been limited to relatively
simple stochastic object models and relatively small image sizes.

In recent
years, motivated by {the theory of} compressive sensing, there has been a paradigm shift in the way that images are formed in computed imaging systems. 
{
Modern sparse reconstruction methods exploit the fact that many objects of interest
can typically be described by use of sparse representations.
Sparse reconstruction} methods have proven to be highly effective at reconstructing images from
under-sampled measurement data in a wide range of medical imaging systems
 including magnetic resonance imaging (MRI) \cite{lustig2007sparse,lustig2008compressed,ramani2010accelerated}
 and X-ray computed tomography (CT)\cite{sidky2008image,bian2010evaluation,stayman2011penalized}.

While a vast amount of research has
 been conducted on the development of sparse reconstruction methods,
 {very little research has addressed the problem of how to optimize
data-acquisition designs for specific diagnostic tasks when sparse reconstruction methods are employed}.
{The minimum amount and quality of measurement data required for estimation
of an image of prescribed accuracy or utility is affected by the sparsity properties
of the object and the specification of the sparse reconstruction method \cite{sidky2008depth}.}
Intuitively, this is to be expected because different sparse reconstruction methods exploit different \textit{a priori}
information about the object. The sparsity-promoting regularization penalties employed by a
reconstruction method can be associated with {SOMs} (i.e., object priors) within a Bayesian
framework.  {However, this seems inconsistent with the conventional notion
\cite{barrett2013foundations,vennart1997icru} that suggests that hardware 
and data-acquisition designs should be optimized by use of an IO and the measurement data, independent
of the choice of image reconstruction method.} 
A potential explanation for this inconsistency is that,
 in modern imaging systems that employ compressive sensing concepts, imaging hardware and sparse
 image reconstruction are innately coupled technologies.

{In this work, a new numerical observer is developed for signal-known-exactly (SKE) detection tasks that can be employed to optimize data-acquisition designs for computed imaging systems when sparse reconstruction methods are employed.
The new numerical observer is referred to as a sparsity-driven observer (SDO). Similar to other approximated IOs, the SDO acts on the measurement data and employs the likelihood ratio as test statistics.
The key difference is that where traditional IOs (and approximations thereof) are predicated upon comprehensive statistical knowledge of the objects imaged, the SDO assumes an SOM that describes only the sparsity properties of the class of objects.}
%
  {In this way, the SDO and sparse reconstruction method will be ``matched" in the sense that they both utilize the same statistical information regarding the class of objects to be imaged.}
  This information is already known to be useful for sparse reconstruction methods.
  As such, the use of the {SDO} will permit task-specific optimization of  data-acquisition
designs with consideration of the (non-linear) sparse reconstruction method to-be-employed. 

The remainder of the article is organized as follows.
In Section II, necessary background {from} signal detection theory and sparse image reconstruction is provided.  
The  {SDO} for the case of i.i.d. Gaussian measurement noise is defined in Section III
 and a semi-analytic procedure for computing it based on variational Bayesian inference for 
sparse linear models is presented.
The validity of the variational approximation employed to compute the {SDO} test statistic 
is investigated in Section IV.
In Section V,
the use of the {SDO} to rank-order data-acquisition designs in a stylized example motivated by
 magnetic resonance imaging (MRI) is demonstrated.
Finally, the article concludes with a discussion of the {SDO} in Section VI.

\section{Background}

\subsection{The {Likelihood Ratio Calculation} for Binary Classification Tasks}
\label{sec:2.1}

Consider a discrete-to-discrete linear imaging model (see sec 7.4 in \cite{barrett2013foundations})
\begin{align}
\label{eqn:model}
\mathbf g=\mathbf H\mathbf f+\mathbf n, 
\end{align}
which {we will consider} to be an accurate approximation of the true continuous-to-discrete tomographic imaging model. 
Here, $\mathbf H:  \mathbb E^{N} \rightarrow \mathbb E^{M} $ denotes the system matrix.
 The vectors $\mathbf g\in \mathbb E^{M}$ and $\mathbf n\in \mathbb E^{M}$ denote the 
discrete measurement data and random noise, respectively.
In this preliminary study,  $\mathbf n$ corresponds to a random vector
whose components are independent and identically distributed (i.i.d.) Gaussian random variables with zero mean and variance $\sigma^2$.
The vector $\mathbf f\in \mathbb E^{N}$ represents a finite-dimensional approximation of the measured object's property distribution.

A signal-known-exactly (SKE) and background-known-statistically (BKS) binary detection task is considered\cite{barrett2013foundations}.  The object vector $\mathbf f$ will be expressed as $\mathbf f = \mathbf f_b +\mathbf f_s$,
where $\mathbf f_s$ denotes the known signal and $\mathbf f_b$ denotes the random background.
  The binary signal{-}detection task requires deciding between the signal absent and signal present hypotheses,
denoted as $\mathcal H_0$ and $\mathcal H_1$:
\begin{align}
\label{eqn:hypothesis}
\mathcal H_0:\,\mathbf g=\mathbf H\mathbf f_b+\mathbf n;\,\,
\mathcal H_1:\,\mathbf g=\mathbf H(\mathbf f_b+\mathbf f_s)+\mathbf n.
\end{align}
Because we have not introduced notation to differentiate random from non-random vectors, it should be emphasized that,
in the SKE-BKS detection task,
$\mathbf f_b$, $\mathbf n$, and $\mathbf g$ are random vectors, while $\mathbf f_s$ is non-random.


To make a decision, the IO employs a test statistic corresponding to the likelihood ratio: 
\begin{align}
\Lambda(\mathbf g) = \frac{P(\mathbf g|\mathcal H_1)}{P(\mathbf g|\mathcal H_0)},
\end{align}
where $P(\mathbf g|\mathcal H_i)$ denotes the conditional PDF of the data vector under the hypothesis $\mathcal H_i$ ($i=0,1$).
It is well-known that the IO decision strategy is optimal in the sense that 
it yields the best possible receiver operating characteristics (ROC) curve\cite{barrett1998objective,wagner1985unified,vennart1997icru}.

For a SKE/BKS detection task, the likelihood ratio test statistic can be calculated as (see \cite{kupinski2003ideal} and Sec 13.2 in \cite{barrett2013foundations})
\begin{align}
\label{eqn:LR}
\Lambda(\mathbf g) = \int \diff\mathbf f_b\, \Lambda_{BKE}(\mathbf g|\mathbf f_b)P(\mathbf f_b|\mathbf g,\mathcal{H}_0),
\end{align}
where $\Lambda_{BKE}(\mathbf g|\mathbf f_b)$ is the likelihood ratio for the background-known-exactly (BKE) case,
where only the noise vector $\mathbf n$ in Eqs.\ (\ref{eqn:hypothesis}) is random,
 and $P(\mathbf f_b|\mathbf g,\mathcal{H}_0)$ is the posterior PDF that describes $\mathbf f_b$  under hypothesis $\mathcal H_0$.
Note that the posterior PDF that appears in Eq.\ (\ref{eqn:LR}) can be expressed as
\begin{align}\label{eqn:posterior}
P(\mathbf f_b|\mathbf g,\mathcal{H}_0) = \frac{P(\mathbf f_b)P(\mathbf g|\mathbf f_b,\mathcal H_0)}{P(\mathbf g|\mathcal H_0)}=\frac{P(\mathbf f_b)P(\mathbf g|\mathbf f_b,\mathcal H_0)}{\int \diff\mathbf f_bP(\mathbf f_b)P(\mathbf g|\mathbf f_b,\mathcal H_0)},
\end{align}
where $P(\mathbf f_b)$ denotes the stochastic object model
and $P(\mathbf g|\mathbf f_b,\mathcal H_0)$ describes the measurement noise 
under the signal absent hypothesis.

\subsection{Challenges in Computing the {Likelihood Ratio} Test Statistic}

Except for special cases, a closed form solution  for the {likelihood ratio} test statistic given by Eq.\ (\ref{eqn:LR}) is generally unavailable.   
A key challenge  is determination of
 the posterior distribution $P(\mathbf f_b|\mathbf g,\mathcal{H}_0)$.
To analytically compute this posterior distribution via  Eq.\ (\ref{eqn:posterior}),
 closed form expressions for the stochastic object model $P(\mathbf f_b)$ and measurement
noise $P(\mathbf g|\mathbf f_b,\mathcal H_0)$ must be available.
While simple forms for $P(\mathbf g|\mathbf f_b,\mathcal H_0)$ may be available -- such as with the Gaussian
measurement noise model assumed later in this work --  simple forms for
$P(\mathbf f_b)$ are rarely available. 
Additionally, the high-dimensional integrals present in Eqs.\ (\ref{eqn:LR}) and (\ref{eqn:posterior})
may be difficult to compute. 
When analytical formula are not available for the posterior distribution,
 sampling methods such as the MCMC method can be employed\cite{kupinski2003ideal}. 
However,  this method can be computationally intensive and can require expertise to run properly {as noted in the introduction}.

\subsection{Maximum a Posteriori (MAP) Estimation Using Sparsity-Promoting Priors}

Over the last decade, sparsity has emerged as a
widely employed concept in signal processing and image
 reconstruction\cite{elad2010role,lustig2007sparse,stayman2011penalized,bian2010evaluation}. 
 A vector $\mathbf f$ is
said to be sparse if most of its components are equal to zero.
 Medical images are, in general, not strictly
sparse, but often can be accurately approximated as such in an appropriate transform domain. Expansion
coefficients that represent an image in linear transform spaces typically have super-Gaussian distributions (peaked,
heavy-tailed) \cite{starck2010sparse}.


Sparse reconstruction algorithms employ a simple and well-known statistical model to describe object sparseness.
 Let $\mathbf B\in \mathbb E^N\rightarrow\mathbb R^Q$ be {a sparsifying transform} that maps an object vector $\mathbf f\in\mathbb E^N$ into a transform space as
\begin{align}
\label{eqn:sparsetransform}
\mathbf w = \mathbf B\mathbf f\in\mathbb R^Q,
\end{align}
where $\mathbf w$ is the transformed vector that is designed
 to be approximately sparse. {The dimension $Q$ of the transformed space depends on the sparsifying matrix chosen.}
 Commonly employed choices for $\mathbf B$ in the image processing literature include wavelet transforms and discrete first derivatives. {The statistical properties of the object that
 are related to sparsity} can be described by an i.i.d. Laplacian distribution\cite{starck2010sparse}:
\begin{align}
\label{eqn:prior}
P(\mathbf f) = \frac\tau2\prod_{i=1}^Q\exp(-\tau| w_i|).
\end{align}
Here, $ w_i$ denotes the $i$-th component of $\mathbf w$ and $\tau\geq0$ is a scale parameter of
the Laplacian distribution. The value of $\tau$ determines how peaked about the origin the distribution is,
 which, in effect, reflects the sparsity of $\mathbf w$. For a given ensemble of objects, this parameter can be empirically determined as described in Appendix\ \ref{app:tau}. For a suitable choice of $\mathbf B$, Eq.\ (\ref{eqn:prior})
 will represent a sparse statistical prior. 

A sparse statistical prior can be utilized to regularize the image reconstruction problem
when maximum a posteriori (MAP) estimation is employed.
A MAP estimate of $\mathbf f$ is defined as

\begin{align}
\label{eqn:MAP1}
\mathbf {\hat f} = \arg\underset{\mathbf f}{\max}\, P(\mathbf f|\mathbf g) = -\arg\underset{\mathbf f}{\min}\, \log \left[P(\mathbf g|\mathbf f)P(\mathbf f)\right].
\end{align}
For the special case of i.i.d. Gaussian noise considered below,
the likelihood function $P(\mathbf g|\mathbf f)$ is given by
\begin{align}
\label{eqn:likelihood}
P(\mathbf g|\mathbf f) &=\mathcal N(\mathbf g|\mathbf H\mathbf f,\sigma^2\mathbf I) \nonumber\\
&=  (2\pi\sigma^2)^{-M/2}\exp\left(-{||\mathbf g-\mathbf H\mathbf f||^2}/{2\sigma^2}\right),
\end{align}
where $\sigma^2$ denotes the noise variance that is assumed to be object-independent and
$\mathbf I$ is the $M\times M$ identity matrix. {Generalization to a more complex noise model is discussed in Sec.~\ref{sec:discussion}.}
Based on Eqs.\ (\ref{eqn:prior}) and (\ref{eqn:likelihood}), the MAP estimate can be
computed as 
\begin{align}
\label{eqn:MAP2}
\mathbf {\hat f} = \arg\underset{\mathbf f}{\min}\, \left({||\mathbf g-\mathbf H\mathbf f||^2}+2\sigma^2\tau\sum_{i=1}^Q|w_i|\right).
\end{align}
Equation\ (\ref{eqn:MAP2})
describes a penalized least squares (PLS) estimate of $\mathbf f$ that employs an $l_1$-norm-based penalty term.
This demonstrates the connection between sparse object priors
 of the form given in Eq.\ (\ref{eqn:prior}) and commonly employed sparse image reconstruction methods.

\section{Computation of the {Sparsity-Driven Observer} Test Statistic}
\label{sec:teststatistic}
Below, the test statistic for the {SDO} is defined
for the case of independent identically distributed (i.i.d.) Gaussian measurement noise.  
A semi-analytic procedure for computing it based on variational Bayesian inference for
sparse linear models is presented subsequently.


\subsection{Definition of the {SDO} Test Statistic}

Formally, the {SDO} test statistic is defined by Eq.\ (\ref{eqn:LR}) when the stochastic  object model  $P(\mathbf f_b)$
that appears in Eq.\ (\ref{eqn:posterior})
 is specified by a sparse prior of the form given in Eq.\ (\ref{eqn:prior}).
In this way, the {SDO} and sparse reconstruction method described by Eq.\ (\ref{eqn:MAP2}) will be matched in the sense that they
both assume the same statistical information about the class of objects to be imaged. 

In this work, we will assume that the components of the noise vector $\mathbf n$ 
are i.i.d. Gaussian random variables having zero mean and variance $\sigma^2$. 
{Thus},
  $\Lambda_{BKE}(\mathbf g|\mathbf f_b)$
 is given by
\begin{align}
\label{eqn:BKE}
&\Lambda_{BKE}(\mathbf g|\mathbf f_b) = \frac{\mathcal N(\mathbf g|\mathbf H(\mathbf f_b+\mathbf f_s),\sigma^2\mathbf I)}{\mathcal N(\mathbf g|\mathbf H\mathbf f_b,\sigma^2\mathbf I)}\nonumber\\
&= \exp\left[({2\mathbf g^\dagger \mathbf H\mathbf f_s-\|\mathbf H\mathbf f_b+\mathbf H\mathbf f_s\|^2+\|\mathbf H\mathbf f_b\|^2})/{2\sigma^2}\right],
\end{align} 
where $\dagger$ denotes the adjoint.

In order to compute the {SDO} test statistic according to Eq.\ (\ref{eqn:LR}), the posterior distribution  $P(\mathbf f_b|\mathbf g,\mathcal{H}_0)$
needs to be determined
for the case where the sparse prior in Eq.\ (\ref{eqn:prior}) and noise model in 
Eq.\ (\ref{eqn:likelihood})  are assumed.
To accomplish this, methods for variational Bayesian inference are employed as described next.

\subsection{Estimation of $P(\mathbf f_b|\mathbf g,\mathcal{H}_0)$ by use of a Variational Approximation}
\label{subsec:variational_approximation}

To circumvent the aforementioned challenges associated with determining $P(\mathbf f_b|\mathbf g,\mathcal{H}_0)$,
 computational tools developed for variational Bayesian inference with large-scale
 sparse linear models {were} employed\cite{seeger2008bayesian,seeger2011large}.
 Variational inference methods employ an analytical approximation to the posterior probability
distribution, and represent alternatives to
 computationally cumbersome MCMC methods
 for inference over complicated distributions that are difficult to evaluate or sample.
 Whereas the MCMC methods provide a numerical
approximation to the exact posterior by use of a set of samples, variational Bayesian methods provide a
locally optimal, exact analytical solution to an approximation of the posterior distribution\cite{fox2012tutorial,tzikas2008variational,seeger2010variational}.
Moreover, {compared with the MCMC methods, 
 variational methods have better scalability as
 the size of the problem increases \cite{kupinski2003ideal} and
can potentially reduce computation times by several orders of magnitude}.
%

By use of the sparse prior in Eq.\ (\ref{eqn:prior}) and likelihood function
in Eq.\ (\ref{eqn:likelihood}),
 the posterior distribution can be expressed as 
\begin{align}
\label{eqn:posterior2}
P(\mathbf f_b|\mathbf g,\mathcal{H}_0)_{P(\mathbf f_b)} =Z^{-1}\mathcal N(\mathbf g|\mathbf H\mathbf f_b,\sigma^2\mathbf I)\prod_{i=1}^Q \exp(-\tau| w_i|),
\end{align}
with
\begin{align}
Z = \int \diff\mathbf f_b\mathcal N(\mathbf g|\mathbf H\mathbf f_b,\sigma^2\mathbf I)\prod_{i=1}^Q \exp(-\tau| w_i|).
\end{align}
Here, the notation $P(\mathbf f_b|\mathbf g,\mathcal H_0)_{P(\mathbf f_b)}$ indicates that
this is  the posterior distribution corresponding to the prior $P(\mathbf f_b)$. 
 The variational sparse Bayesian inference method proposed
by Seeger and colleagues\cite{seeger2008bayesian,seeger2011large} can be used to approximate the
true posterior distribution by a Gaussian distribution where the marginal variances are determined by use
of the measured data $\mathbf g$ as described below.
To accomplish this, an approximation to the Laplacian prior can be employed.
 It has been demonstrated\cite{seeger2008bayesian,seeger2011large} that{, based on Cauchy inequality,} the
sparse prior in Eq.\ (\ref{eqn:prior}) can be lower bounded by a {parameterized} Gaussian distribution of the form 
\begin{align}
\label{eqn:approximateprior}
\tilde P(\mathbf f_b;\boldsymbol{\gamma}) = \prod_{i=1}^{Q}\frac\tau2\exp\left(-\frac{\tau^2\gamma_i}{2}\right)\exp\left(-\frac{w_i^2}{2\gamma_i}\right),
\end{align}
where $\boldsymbol{\gamma} = [\gamma_1, \gamma_2,...,\gamma_Q]$, with {$\gamma_i>0$} ($i=1,2,...,Q$)
denoting parameters that {define the shape of the distribution. These parameters }specify the marginal variances of the Gaussian distribution that
will be determined by use of the measured data $\mathbf g$. {Note that the distribution is not normalized, however, this does not affect the posterior sought because the normalization constant ultimately cancels out.}

Based on this approximation of the prior distribution, an approximation of the
 posterior distribution can be expressed as 
{
\begin{align}
\label{eqn:a_posterior}
P(\mathbf f_b|\mathbf g,\mathcal{H}_0)_{\tilde P(\mathbf f_b;\boldsymbol{\gamma})} =
Z_a^{-1}\mathcal N(\mathbf g|\mathbf H\mathbf f_b,\sigma^2\mathbf I)\exp\left(-{\mathbf w^T \boldsymbol{\Gamma}^{-1}\mathbf w}/2\right),
\end{align}
where
\begin{align}
Z_a = \int \diff\mathbf f_b\,
&\mathcal N(\mathbf g|\mathbf H\mathbf f_b,\sigma^2\mathbf I)\exp\left(-{\mathbf w^T\boldsymbol{\Gamma}^{-1}\mathbf w}/2\right).
\end{align}
}
Here, $\boldsymbol{\Gamma}$ is a $Q\times Q$ diagonal matrix whose diagonal elements are specified by the components of the $\boldsymbol\gamma$ and $P(\mathbf f_b|\mathbf g,\mathcal H_0)_{\tilde P(\mathbf f_b;\boldsymbol{\gamma})}$ denotes
 the posterior distribution corresponding to the approximated prior $\tilde P(\mathbf f_b;\boldsymbol\gamma)$ defined 
in Eq.\ (\ref{eqn:approximateprior}).
In this work, as shown below, the approximate posterior $P(\mathbf f_b|\mathbf g,\mathcal{H}_0)_{\tilde P(\mathbf f_b;\boldsymbol{\gamma})}$ will be useful
because it can be employed to semi-analytically compute the {SDO} test statistic.

Finally, the parameter vector $\boldsymbol \gamma$ 
needs to be determined in a way that renders  $P(\mathbf f_b|\mathbf g,\mathcal{H}_0)_{\tilde P(\mathbf f_b;\boldsymbol \gamma)}$ a close approximation of the true posterior distribution
$P(\mathbf f_b|\mathbf g,\mathcal{H}_0)_{P(\mathbf f_b)}$.
This can be accomplished by determining $\boldsymbol \gamma$ such that the Kullback-Leibler (KL) divergence between the two distributions is minimized:
\begin{equation}\label{eqn:gamma}
	\boldsymbol{\hat\gamma}(\mathbf g) = \arg\underset{\boldsymbol{\gamma}}{\min}\, KL[P(\mathbf f_b|\mathbf g,\mathcal H_0)_{P(\mathbf f_b)} || P(\mathbf f_b|\mathbf g,\mathcal H_0)_{\tilde P(\mathbf f_b;\boldsymbol \gamma)}], 
\end{equation}
where $KL[a || b]$ denotes the KL divergence from $b$ to $a$ \cite{kullback1951information}. 
The matrix $\boldsymbol{\hat\Gamma}(\mathbf g)$ denotes the corresponding
estimate of $\boldsymbol{\Gamma}$, 
whose diagonal elements are specified by the components of $\boldsymbol{\hat\gamma}(\mathbf g)$.
It is important to note that the estimated parameter vector $\boldsymbol{\hat \gamma}$, and hence the explicit
form of the approximated posterior distribution $P(\mathbf f_b|\mathbf g,\mathcal H_0)_{\tilde P(\mathbf f_b;\boldsymbol{\hat\gamma})}$,
depends on the measured data $\mathbf g$. As such, this variational approach is not equivalent
to simply replacing the sparse prior by a Gaussian one. This fact is key to the effectiveness of the approach.

Fortunately, tools from convex optimization can be employed to solve Eq.\ (\ref{eqn:gamma}).
Specifically, it has
been established that Eq.\ (\ref{eqn:gamma}) corresponds to a convex optimization problem that can be efficiently solved
by use of a double-loop optimization algorithm\cite{seeger2009speeding,seeger2010gaussian,seeger2010optimization}.
 Details of the algorithm are provided in the Appendix \ref{app:A}.

\subsection{Calculation of the {SDO} Test Statistic}


An estimate of the {SDO} test statistic can be obtained by use of Eq.\ (\ref{eqn:LR}) when
$P(\mathbf f_b|\mathbf g,\mathcal{H}_0)$ is replaced by $P(\mathbf f_b|\mathbf g,\mathcal{H}_0)_{\tilde P(\mathbf f_b;\boldsymbol{\hat\gamma})}$.
In this case, the integral in Eq.\ (\ref{eqn:LR}) can be analytically computed and the test statistic is given by
\begin{equation}
\label{eqn:LR2}
\Lambda(\mathbf g) = \exp\left[ (\mathbf g - \frac12 \mathbf H\mathbf f_s)^\dagger \boldsymbol\Sigma_n^{-1}(\mathbf H\mathbf f_s-\mathbf H \boldsymbol\Sigma(\mathbf g) \mathbf H^\dagger\boldsymbol\Sigma_n^{-1}\mathbf H\mathbf f_s) \right],
\end{equation}
where {
\begin{equation}
\label{eqn:Sigma}
\boldsymbol\Sigma(\mathbf g) = \{\mathbf H^\dagger\boldsymbol\Sigma_n^{-1}\mathbf H+\mathbf B^\dagger[\boldsymbol{\hat\Gamma}(\mathbf g)]^{-1}\mathbf B\}^{-1},
\end{equation}
}
$\boldsymbol\Sigma_n = \sigma^2\boldsymbol I$ is the noise covariance matrix,  
and $\mathbf B$ is defined as in Eq.\ (\ref{eqn:sparsetransform}).\footnote{{The derivation of Eq.\ (\ref{eqn:LR2}) is provided in Appendix \ref{app:D}.}}

Note that $\boldsymbol\Sigma(\mathbf g)$ is defined as a matrix inversion that can be challenging to evaluate. 
 Thus, instead of directly calculating $\mathbf {\hat f}_s = \boldsymbol\Sigma(\mathbf g)\mathbf H^\dagger \boldsymbol\Sigma_n^{-1}\mathbf H \mathbf f_s$, a system
of linear equations 
\begin{equation}
\label{eqn:LRder1}
[\boldsymbol\Sigma(\mathbf g)]^{-1}\mathbf {\hat f}_s = \mathbf H^\dagger \boldsymbol\Sigma_n^{-1}\mathbf H \mathbf f_s 
\end{equation}
can be solved for $\mathbf {\hat f}_s$. Because $[\boldsymbol\Sigma(\mathbf g)]^{-1}$ has an explicit analytical formula {as provided by Eq.\ (\ref{eqn:Sigma})} and is {positive-definite}, this
system of linear equations can be solved efficiently by use of a linear conjugate gradient (CG) algorithm. {Substitution from Eq.\ (\ref{eqn:Sigma}) into Eq. \ (\ref{eqn:LRder1}) yields}
\begin{equation}
\label{eqn:LRder2}
\mathbf H\mathbf H^\dagger (\mathbf {\hat f}_s-\mathbf f_s) - \mathbf B^\dagger[\boldsymbol\Sigma_n^{-1}\boldsymbol{\hat\Gamma}(\mathbf g)]^{-1}\mathbf B \mathbf {\hat f}_s = 0.
\end{equation}
 The solution to this equation can be computed as
\begin{equation}
\label{eqn:LRder3}
\mathbf{\hat f}_s = \arg\underset{\mathbf x}{\min} \, \frac12(\mathbf x-\mathbf f_s)^\dagger\mathbf H^\dagger\mathbf H(\mathbf x-\mathbf f_s)+\frac12\mathbf x^\dagger \mathbf B^\dagger [\boldsymbol\Sigma_n^{-1}\boldsymbol{\hat\Gamma}(\mathbf g)]^{-1} \mathbf B\mathbf x.
\end{equation}
In this way, evaluation of Eq.\ (\ref{eqn:LR2}) involves only matrix-vector multiplications related to $\mathbf B$, $\mathbf H$ and the adjoints thereof, which are routinely employed in iterative image reconstruction algorithms and can be calculated efficiently on-the-fly. 

In summary, the {SDO} test statistic can be computed as\cite{wang2015sparsity}:
\begin{equation}
\label{eqn:LR3}
\Lambda(\mathbf g) = \exp\left[ (\mathbf g - \frac12 \mathbf H\mathbf f_s)^\dagger \boldsymbol \Sigma_n^{-1}\mathbf H (\mathbf f_s-\mathbf {\hat f}_s) \right],
\end{equation}
where
\begin{equation}
\label{eqn:LR3-2}
\mathbf {\hat f}_s = \arg\underset{\mathbf x}{\min} \, ||\mathbf H\mathbf x-\mathbf H\mathbf f_s||^2+||[\boldsymbol\Sigma_n^{-1}\boldsymbol{\hat\Gamma}(\mathbf g)]^{-1/2}\mathbf B\mathbf x||^2.
\end{equation}

\section{Investigation of the Accuracy of the Variational Approximation}

While the variational approximation utilized to establish the approximate posterior distribution in Eq.\ (\ref{eqn:a_posterior}) is useful
in the sense that it permits semi-analytic calculation of the {SDO} test statistic,
it is important to ensure that it can produce an accurate estimate of the posterior distribution
 corresponding to the sparse prior specified by Eq.\ (\ref{eqn:prior}).
Quantification of the accuracy of results obtained via variational Bayesian methods can be difficult
and is not routinely reported in the literature. {Some validation studies have been performed to evaluate the accuracy of the approximation in a more general context\cite{nickisch2010bayesian}. In this section, two specially designed studies are 
presented to suggest that the variational approximation employed in Sec.\ \ref{sec:teststatistic}
 can yield accurate estimates of the posterior distribution and thus the {SDO} test statistic in this application.}

\subsection{Stylized One-Dimensional Problem}

{
  It is generally difficult to evaluate the
 accuracy of a variational approximation in a high dimensional space.
 Below, a one-dimensional (1D) stylized problem corresponding to a Laplacian prior was considered first. }By design, 
an analytical form for the exact posterior distribution of this problem exists. 
The exact posterior distribution was compared directly to the approximate
posterior distribution produced by use of the variational method in Sec.\ \ref{sec:teststatistic}-B.


Consider a 1D model of the form
\begin{equation}
\label{eqn:1d:model}
y = x+\tilde n,
\end{equation}
where the random variable  $y$ describes the measurement, 
the random variable $x$ describes the object property sought,
 and the random variable $\tilde n$ describes Gaussian measurement noise 
with zero mean and variance $\sigma^2$.
 The variable $x$ is assumed to follow a Laplacian distribution:
\begin{equation}
\label{eqn:1d:x}
P(x) =\frac{\tau}2 \exp(-\tau|x|).
\end{equation}
Given a measured value $y$, the true posterior distribution and the approximated posterior distribution obtained by use of the variational Bayesian method will be compared. 

\subsubsection{Posterior distributions}


The true posterior distribution can be calculated analytically in this simple 1D problem:
\begin{align}
\label{eqn:1d:post}
P(x|y) = \frac{2}{\sigma\sqrt{2\pi}}{\exp\left(-\tau|x|\right)\exp\left(-{(y-x)^2}/({2\sigma^2})\right)}/C,
\end{align}
where 
\begin{align}
\label{eqn:1d:postC}
C = &\exp\left(({\tau^2\sigma^2-2\tau y})/{2}\right)\left(1+\text{erf}({-\mu_1}/({\sqrt{2}\sigma}))\right)\nonumber\\
&+\exp\left(({\tau^2\sigma^2+2\tau y})/{2}\right)\left(1+\text{erf}({-\mu_2}/({\sqrt{2}\sigma}))\right).
\end{align}
and erf$()$ is the Gauss error function. 
The variational Bayesian method approximates the prior distribution by a Gaussian distribution as 
\begin{align}
\label{eqn:1d:ax}
\tilde P(x) =\frac{\tau}2 \exp\left(-\frac{\tau^2\gamma}{2}\right)\exp\left(-\frac{x^2}{2\gamma}\right),
\end{align}
where $\gamma$ is the variational parameter to be determined. Subsequently, an approximate PDF describing $y$ can be calculated as
\begin{align}
\label{eqn:1d:apost}
\tilde P(x|y) &= \frac1\sigma\sqrt{\frac{1+\sigma^2/\gamma}{2\pi}}\\
&\times\exp\left\{-\frac1{2\sigma^2}\left[(1+\sigma^2/\gamma)x^2-2xy+\frac{y^2}{1+\sigma^2/\gamma}\right]\right\}.\nonumber
\end{align}
The variational Bayesian inference method minimizes the KL-divergence between the two posterior distributions with respect to $\gamma$. The process can be proven to be equivalent to maximizing\cite{nickisch2010bayesian}
\begin{align}
\label{eqn:1d:ay}
\tilde P(y) = &\frac{\tau}{2\sqrt{1+\sigma^2/\gamma}}\nonumber\\
&\times\exp\left(-\frac{\tau^2\gamma}{2}-\frac{y^2}{2\sigma^2}+\frac{y^2}{2\sigma^2(1+\sigma^2/\gamma)}\right)
\end{align}
with respect to $\gamma$, given the observed value of $y$.
The optimization problem in this 1D case can be solved by the simplex search method\cite{lagarias1998convergence}. The approximated posterior distribution can be determined based on the estimated $\gamma$. 

\subsubsection{Unbiased observation}

\begin{figure}[h!]
\centering
\includegraphics[width=\linewidth]{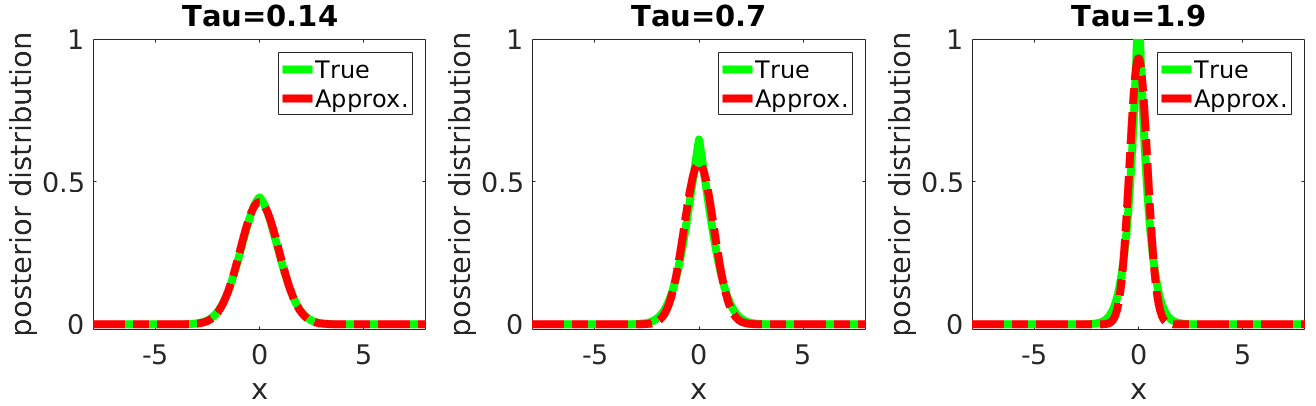}

\caption{ The true posterior distribution and the approximated posterior distribution with different $\tau$ values. The KL-divergences are 0.0006, 0.014, 0.070 respectively.}
 \label{fig:1d:unbiased}
\end{figure}

First, an unbiased scenario was investigated in which the observation $y$ is exactly the same as the most likely $x$ value. Given the Laplacian distribution of $x$, an unbiased observation requires $y=0$. The standard deviation for the Gaussian noise was $\sigma=1$. The true posterior distributions and the approximated posterior distributions corresponding to three different prior Laplacian distributions are plotted in Fig.~\ref{fig:1d:unbiased}, with the $\tau$ value being $0.14,0.7$, and $1.9$ respectively. It can be seen in Fig.~\ref{fig:1d:unbiased} that the true and approximated posterior distributions are close to each other {in terms of the shape of the distribution}.

Furthermore, the KL-divergence was employed to quantitatively measure the distance between the true and approximated posterior distribution. The KL-divergences for the three cases are $0.0006,0.014$, and $0.070$ respectively. It was observed that the approximation is more accurate if the signal is less sparse ($\tau$  is smaller) {as expected}.


\subsubsection{Biased observation}

\begin{figure}[h!]
\centering
\includegraphics[width=\linewidth]{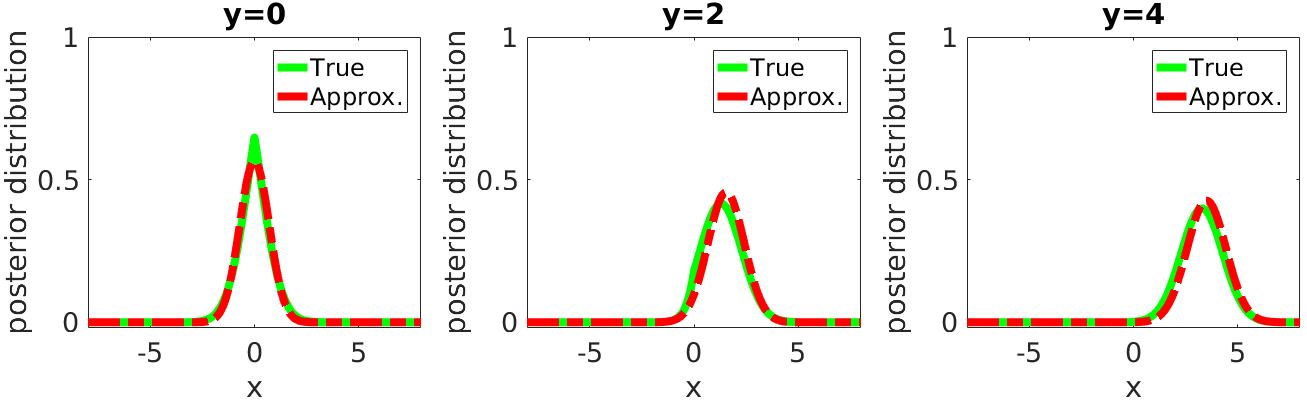}

\caption{ The true posterior distribution and the approximated posterior distribution with different observations. The KL-divergences are 0.014, 0.026, 0.031 respectively.}
 \label{fig:1d:biased}
\end{figure}

A biased scenario was also investigated in which the observation $y$ is different from {0, which is the most likely value for $x$}. In this study, we changed the observation $y$ to be $2$ and $4$, which reflects that the observation was affected by the signal randomness and measurement noise and thus became biased. The parameters for the Laplacian distribution and the Gaussian noise distribution were $\tau=0.7$ and $\sigma=1$. The true posterior distributions and the approximated posterior distributions corresponding to the {two measurement $y$ values as well as the unbiased observation of $y=0$} are plotted in Fig.~\ref{fig:1d:biased}, from which we can see that the true and approximated posterior distributions have similar peak shape and location. The KL-divergences for these three cases are $0.014, 0.026$, and $0.031$ respectively, indicating that the approximation is more accurate in a less biased observation.

In conclusion, given a Laplacian (sparse) prior assumption, the approximated posterior distribution obtained by use of Eq.\ (\ref{eqn:1d:apost}) was found to be close to the true posterior distribution.

\subsection{Impact of Variational Approximation on Reconstructed Images}
\label{sec:recon-demo}

{When extending the approximation from 1-dimensional space to $N$-dimensional space, it is no longer possible to compare the entire true and approximated distributions.
 Luckily, it is possible to locate and compare the modes of the distributions, which are exactly the MAP estimates. As such, to further investigate the accuracy 
of the variational approximation utilized to establish
 the approximate posterior distribution in Eq.\ (\ref{eqn:a_posterior}), image reconstruction studies were conducted.}  
Images were reconstructed by use of three different PLS estimators.
First, the PLS estimate described in Eq.\ (\ref{eqn:MAP2}) was computed.
  In this case, the $\ell_1$-norm-based penalty term employed by the sparse reconstruction method corresponds to the logarithm
 of the sparse prior given in Eq.\ (\ref{eqn:prior}). The sparsifying transform
in Eq.\ (\ref{eqn:sparsetransform}) was specified as a level-4 Haar wavelet transform {and the parameter $\tau$ was empirically estimated by the method described in Appendix\ \ref{app:tau}}.
Next, a second PLS estimate was computed in which the $\ell_1$-norm-based penalty was
replaced by the logarithm of its variational approximation given in Eq.\ (\ref{eqn:approximateprior}).
 The parameter vector $\boldsymbol{\gamma}$ that specified the Gaussian approximation
 was determined as described in Sec.\ \ref{subsec:variational_approximation}.
These estimates will be referred to as the PLS-$\ell_1$ and
 PLS-$\ell_1-approx$, respectively.
If the variational approximation to the sparse prior is sufficiently accurate,
it can be expected that the PLS-$\ell_1$ and PLS-$\ell_1-approx$ estimates should be similar.
Additionally, a conventional PLS estimate that utilized an $\ell_2$-norm-based penalty term was also
 computed.  These estimates will be referred to as the
PLS-$\ell_1$, PLS-$\ell_1-approx$, and PLS-$\ell_2$, estimates, respectively.


{
\begin{figure}[h!]
\centering
\includegraphics[width=0.7\linewidth]{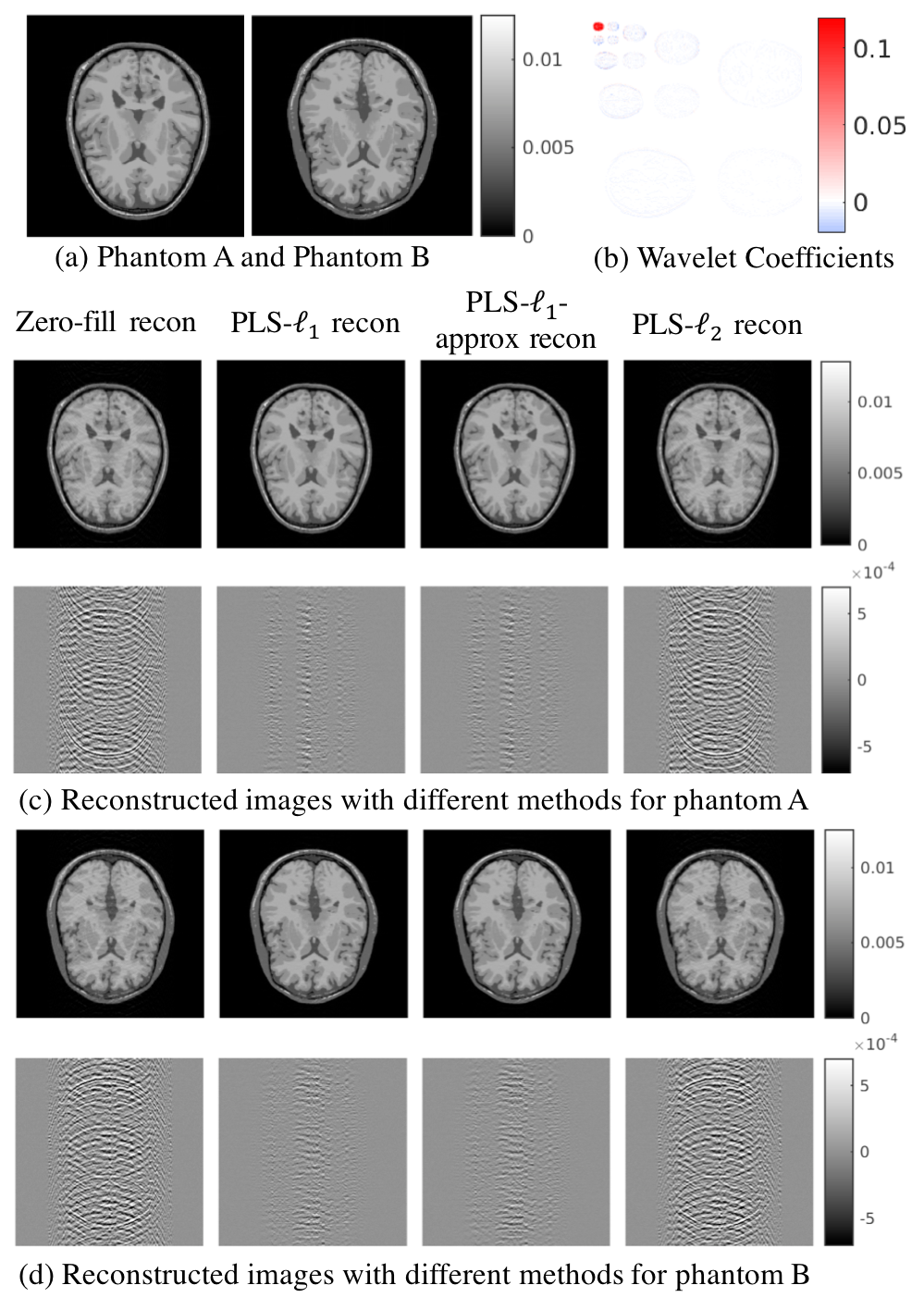}
\caption{{The reconstructed images for two different phantoms (shown in (a)) and different reconstruction methods. Subfigure (b) shows the wavelet coefficients for the first phantom. Zeros are denoted by the white color. The widespread white region in the wavelet image demonstrates the sparseness in the wavelet domain.} Subfigures (c) and (d) show reconstructed images for the two different phantoms. In both (c) and (d), the first row shows the reconstructed images and the second row shows the difference images between the reconstructed images and the true phantom. In each row, from left to right are zero-filling reconstruction, PLS-$\ell_1$ reconstruction, PLS-$\ell_1-approx$ reconstruction, and PLS-$\ell_2$ reconstruction.}
 \label{fig:l1l2}
\end{figure}}

The estimates above were computed from computer-simulated data corresponding
to a stylized two-dimensional (2D) MRI example in which the imaging system sampled random lines from k-space.
Two different  head phantoms  were employed to represent the to-be-imaged object, which
are displayed in {Figs.~\ref{fig:l1l2}-(a)}. {The sparsity of the phantom in the wavelet domain is demonstrated in Figs.~\ref{fig:l1l2}-(b).} 
Additional details regarding these numerical phantoms are described later in Sec.~\ref{sec:sd}.
To demonstrate the impact of under-sampling in the simulated measurement data,
images were reconstructed by  zero-filling the missing k-space regions and applying
  an inverse 2D discrete Fourier transform (DFT). {This method is also known as the pseudoinverse method \cite{lustig2007sparse}.}
The resulting images, shown in the first rows and second columns of Figs.~\ref{fig:l1l2}-(c) and (d),
contain aliasing artifacts.
 The artifacts are clearly visible in the difference images that
are shown in the second rows and first columns of Figs.~\ref{fig:l1l2}-(c) and (d), which were
formed by subtracting the reconstructed images from the corresponding true phantoms.


The images corresponding to the  PLS-$\ell_1$, PLS-$\ell_1-approx$, and PLS-$\ell_2$ estimates
are displayed in the second, third, and fourth columns of the top rows in Figs.~\ref{fig:l1l2}-(c) and (d).
The corresponding difference images are shown in the lower row of each subfigure.
Note that the PLS-$\ell_2$ estimate contains significant artifacts that are similar in nature to those
produced by the zero-filling-inverse DFT approach.  Tuning the regularization parameter 
did not change this observation.
On the other hand,
the PLS-$\ell_1$ and PLS-$\ell_1-approx$ estimates have significantly reduced artifact levels
as compared to the PLS-$\ell_2$ estimate.  This is expected, as sparsity-promoting regularization
methods are known to be more effective for mitigating data incompleteness  than $\ell_2$-based methods.
Moreover, it is observed that the  PLS-$\ell_1$ and PLS-$\ell_1-approx$ image estimates
are nearly indistinguishable. Namely, the magnitude and spatial distribution of errors in these
two image estimates, as reflected in the difference images, are almost the same.
This suggests that the approximated prior in Eq.\ (\ref{eqn:approximateprior}) promotes
  (wavelet-domain)  sparsity
is a manner very similar to how the exact sparse prior Eq.\ (\ref{eqn:prior}) does.
This is evidence  of the accuracy of the variational approximation that underlies the
proposed method for computing the {SDO} test statistic.


\section{System Rank-Ordering Study}

An important application of numerical observers is the  rank-ordering
of imaging system or data-acquisition designs.
 In this section, the {SDO} will be applied to rank four
 different data-acquisition designs via performance in a tumor{-}detection task in a stylized 2D MRI example. 

\subsection{Study Design}
\label{sec:sd}

\begin{figure}[ht!]
\begin{minipage}[t]{0.32\linewidth}
\centering
\includegraphics[width=\linewidth]{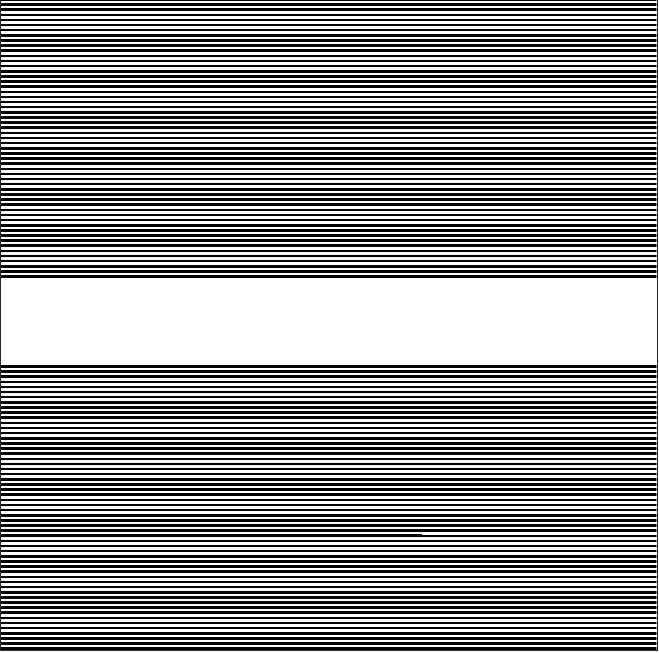}
\subcaption{\label{fig:sys:b} UH}
\end{minipage}
\begin{minipage}[t]{0.36\linewidth}
\centering
\includegraphics[width=0.889\linewidth]{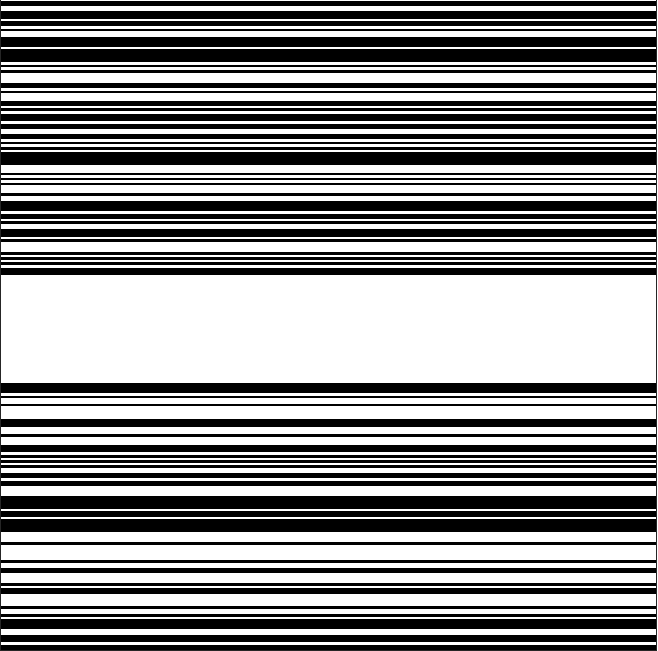}
\subcaption{\label{fig:sys:c} RH}
\end{minipage}%
\begin{minipage}[t]{0.32\linewidth}
\centering
\includegraphics[width=\linewidth]{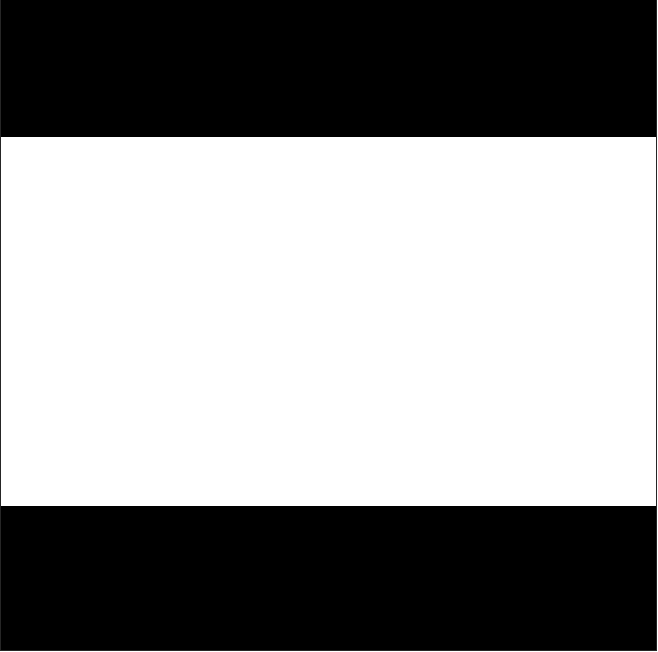}
\subcaption{\label{fig:sys:d} LH}
\end{minipage}
\caption{ Sampling patterns of half-scan candidate data-acquisition designs: 
(a) {UH}: uniform half-scan; 
(b) {RH}: random half-scan; 
(c) {LH}: low-pass half-scan. 
The white lines are the sampled phase-encoding lines in k-space while the black lines are un-sampled. The center of the image corresponds to low-frequency components.}
 \label{fig:sys}
\end{figure}

Four MRI data-acquisition designs were considered for comparison. Each design
 sampled different  256 phase-encoding lines that were
 uniformly-spaced between the highest positive (+$k_{\text{max}}=500$ m$^{-1}$) and negative (-$k_{\text{max}}=-500$ m$^{-1}$) spatial frequencies, as shown in Fig.~\ref{fig:sys}.
 Among the four candidate patterns, the \textbf{full scan (FS)} sampling pattern (not shown) comprises all 256 phase-encoding lines in k-space and serves as a reference. {All three of other patterns} are half-scan schemes
 that each contain 144 sampled phase-encoding lines.
 The \textbf{uniform half-scan (UH)} sampling pattern comprises 72 lines
 within a low-frequency region of k-space  with an 72 additional lines uniformly spaced in the rest of k-space; the \textbf{random half-scan (RH)} sampling pattern comprises the same low-frequency 72 lines
 and 72 additional lines sampled randomly; the \textbf{low-pass half-scan (LH)}
 sampling pattern has all 144 lines in a low-frequency region.

\begin{figure}[h]
\begin{subfigure}{\linewidth}
 \centering
 \includegraphics[width=\linewidth]{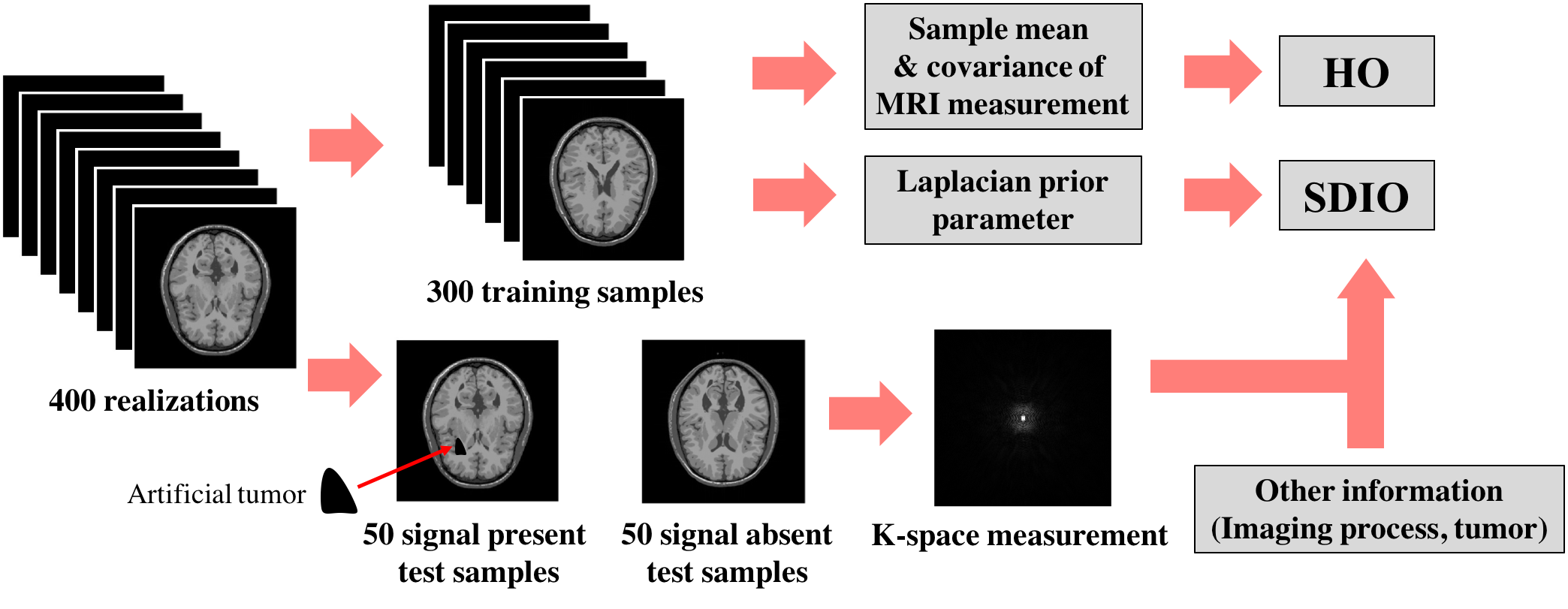}
 \caption{\label{fig:design}}
 \end{subfigure}
 \begin{subfigure}{\linewidth}
  \centering
 \includegraphics[width=\linewidth]{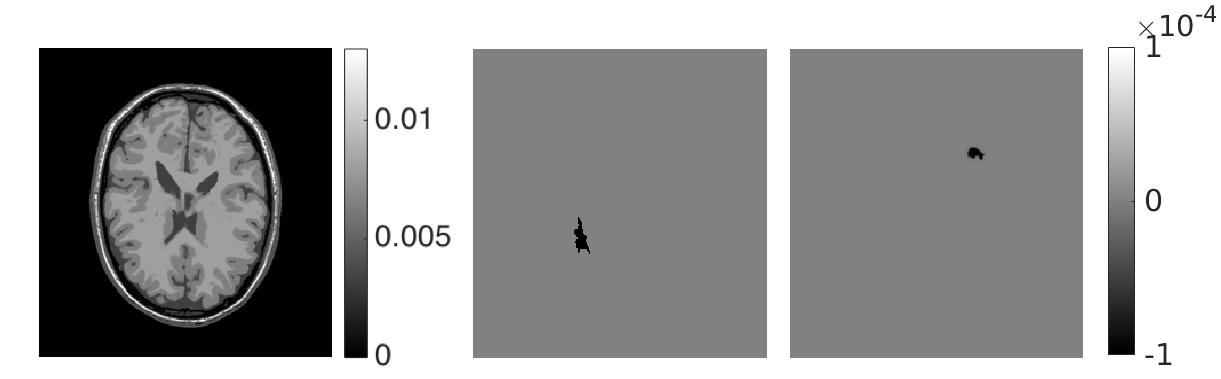}
 \caption{\label{fig:phantom}}
 \end{subfigure}
 \caption{(a) The flowchart of generating the simulated data for the observers. (b) From left to right: a sample background head phantom, tumor phantom 1, tumor phantom 2.}
 \label{fig:recon}
\end{figure}

The process of ranking the candidate designs is illustrated in Fig.~\ref{fig:design}.
Numerical {real-valued} brain phantoms were obtained from an online phantom library\cite{collins1998design}. 
{These phantoms were designed to represent reconstructed T1-weighted brain images while considering some limitations of the imaging process such as the partial volume effect and limited receiver bandwidth.}
This means that a zoomed-in region of the phantom may contain subtle features that resemble artifacts;
however, within the context of our study these will simply represent components of the phantom.
 Out of the available 20 digital 3D brain phantoms, which depicted normal
brain anatomies,  we extracted 400 2D brain slices
 that each contained $256\times256$ pixels with each pixel having a dimension of 1 mm.
 The phantoms were organized into two sets:
 300 phantoms were randomly chosen to form a training set and the remaining
 100 phantoms formed a testing set.
 The samples in the training set were employed to estimate the first and second order statistics required in the Hotelling observer (HO) and the Laplacian parameter $\tau$ for {SDO}.
 The 100 testing images were further divided into 50 signal
 present test images and 50 signal absent test images to assess tumor{-}detection performance for both the {SDO} and the HO. 
To produce the 50 signal present images, a simulated tumor phantom (tumor phantom 1) was generated (Fig.\ \ref{fig:phantom}, middle panel) and superimposed on a slice of the brain phantom.
Idealized k-space measurement data were produced corresponding to the four sampling patterns.
Zero-mean  i.i.d.\ {complex} Gaussian noise of standard deviation $\sigma=2\times 10^{-5}$ was added to the measurements to create noisy data sets{, generated by creating real-valued Gaussian noise with a standard deviation of $\sigma'=\sigma/\sqrt{2}$ for the real and imaginary parts separately.} This noise level corresponds to 20\% of the maximum value of the tumor signal.
Finally, the {SDO} test statistic was computed by use of the k-space data corresponding to 
each phantom in the testing set and ROC curves were generated. {The implementation details of the {SDO} test statistical calculation are given in Appendix \ref{app:A}. The Hotelling observer implementation follows a
standard procedure\cite{wang2015sparsity}.}
All ROC curves were fit based on the semi-parametric binormal model\cite{metz1980statistical}.
 The area under the curves (AUC) were computed as a figure-of-merit for signal
detection performance. 
The procedure above was repeated by use of second tumor phantom (tumor phantom 2), as shown in Fig.\ \ref{fig:phantom}, right panel.
The contrast of this signal was the same as that of the first signal, with only the size, shape, and 
location of the signal being different.
In this second study, the training and testing sets were re-formed by random sampling.



The effect of the training set size on the performance of
the numerical observers was considered. In the implementation of the {SDO}, the sparsifying transform $\mathbf B$ was selected to be a four-level Haar wavelet transform. {Haar wavelet is the simplest wavelet basis and four-level decomposition is a fair choice for this study where the discretized object size is $256\times256$.} In this case, {$\mathbf B$ is a square matrix and thus the dimension of the transformed linear space $Q=N=65536$}, which is the dimension of the object vector $\mathbf f$. The parameter $\tau$ was 
estimated by use of the training samples {as described in Appendix\ \ref{app:tau}}.
 It was observed that changing the number of training samples from 1 to 300
  lead to a maximum fluctuation of only $0.01$\% in the estimated $\tau$ value.
 This suggests that, in the case considered,  the number of training samples
 did not significantly change the estimated $\tau$ value and hence did
 not affect strongly the {SDO} performance.
 Consequently, here, $\tau$ was estimated from only one training sample.
 In other situations {when the training set comes from multiple sources and has larger ensemble variation}, multiple training samples may be necessary to accurately
estimate $\tau$; however, as long as the training images are {on the same order of magnitude}, the required number of training samples remains low because only one parameter needs to be estimated. 
On the other hand, the number of training samples employed to estimate the
background covariance matrix can significantly affect HO performance.
In this study, all 300 training samples were utilized for this purpose. For a comparison, a second, limited data situation was simulated by training the HO by use of only 100 training samples.




\subsection{Rank-Ordering of Data Acquisition Designs}

\begin{figure*}[h]
\centering
\includegraphics[width=\linewidth]{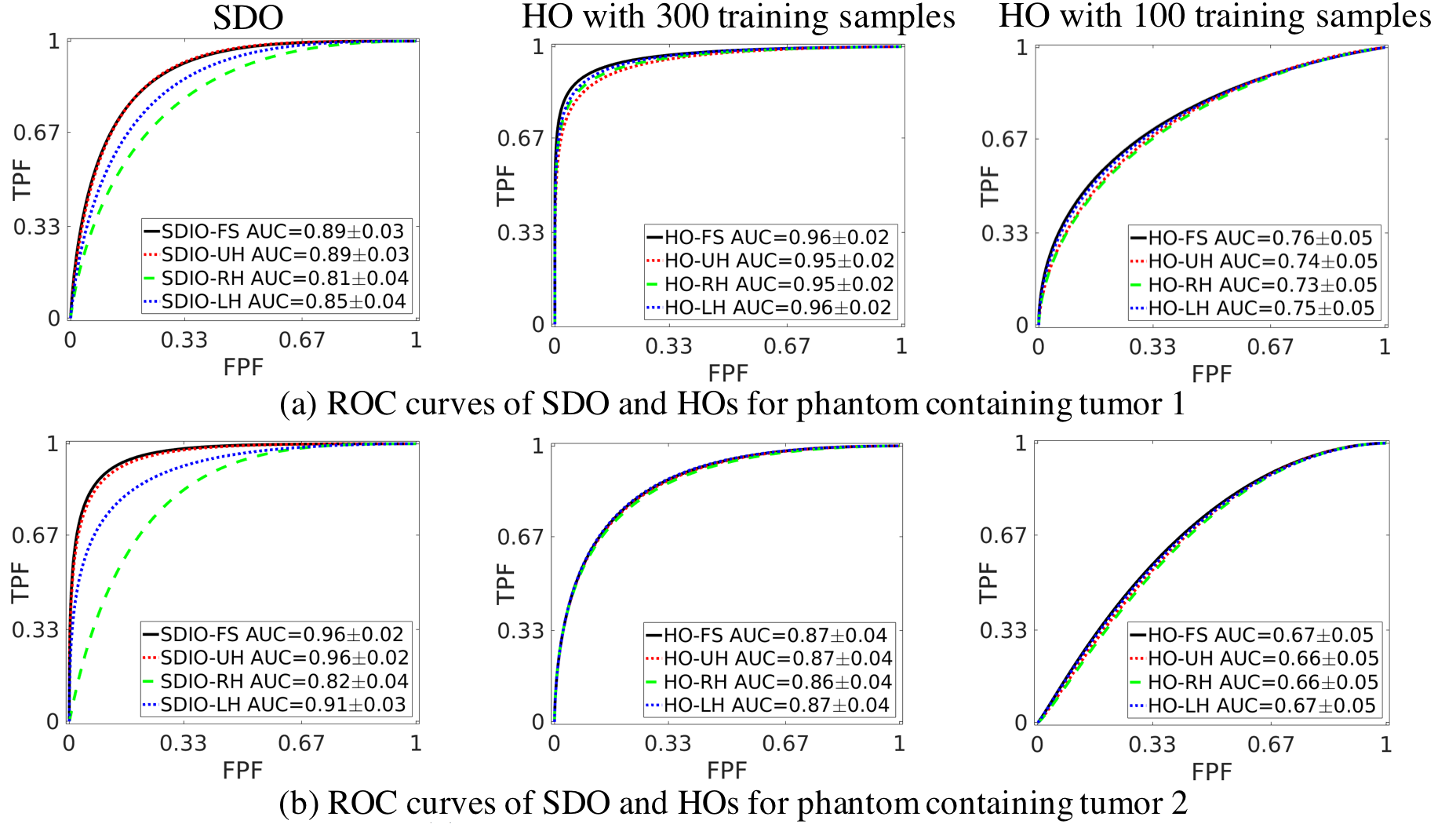}

%
\caption{ ROC curves and AUC values for different data-acquisition designs.
Subfigures (a) and (b) correspond to use of phantoms that contain tumors 1 and 2.
In each subfigure, the numerical observers employed were the  {SDO} (left panel),
 HO with 300 training images (middle panel), and HO with 100 training images (right panel). 
The curves corresponding to different colors correspond to the different designs:
{black: full scan (FS); red: uniform half-scan (UH); green: random half-scan (RH); blue: low-pass half-scan (LH).}
The {SDO} yields a larger separation between ROC curves than does the HO, which
 facilitates their rank ordering.}
 \label{fig:roc}
\end{figure*}


The ROC curves generated for each of the four candidate data-acquisition
designs are shown in Fig.~\ref{fig:roc}. Subfigures (a) and (b) correspond to
the use of tumor phantom 1 and tumor phantom 2, as described above.
As shown in Fig.~\ref{fig:roc}a, the half-scan ROC curves that describe the {SDO} performance
are separated and provide a distinguishable ranking according to the AUC values: \textbf{FS} $\approx$ \textbf{UH} $>$ \textbf{LH} $>$ \textbf{RH}.
 On the other hand, the ROC curves that describe the HO performance are quite close to each other,
and not separated as much as the curves for the {SDO}.
Moreover, the AUC values describing HO performance suggest a different ranking of
half-scan designs: \textbf{FS} $>$ \textbf{LH} $>$ \textbf{UH} $>$ \textbf{RH} (HO with 100 training samples) or \textbf{FS} $>$ \textbf{LH} $>$ \textbf{RH} $>$ \textbf{UH} (HO with 300 training samples). 
The ROC curves in Fig.~\ref{fig:roc}b yield similar observations.
However, in that case, the four ROC curves corresponding to the HO appear nearly indistinguishable, so no clear
rank-ordering can be achieved.


In summary, it is expected that the FS data-acquisition design should yield the best signal{-}detection performance, and the results produced by use of both the {SDO} and HO confirm this.
When restricted to a half-scan design, 
 the {SDO} ranks the the {UH} as best 
 while the HO ranks the {LH} as best (when it is still capable of producing a ranking). 
This demonstrates that the {SDO}{, which utilizes a reconstruction prior,} and HO{, which ignores image reconstruction entirely,} can produce different rank orderings of designs.

\subsection{Visual Inspection of Images}

\begin{figure}[h]
\centering
\includegraphics[width=0.7\linewidth]{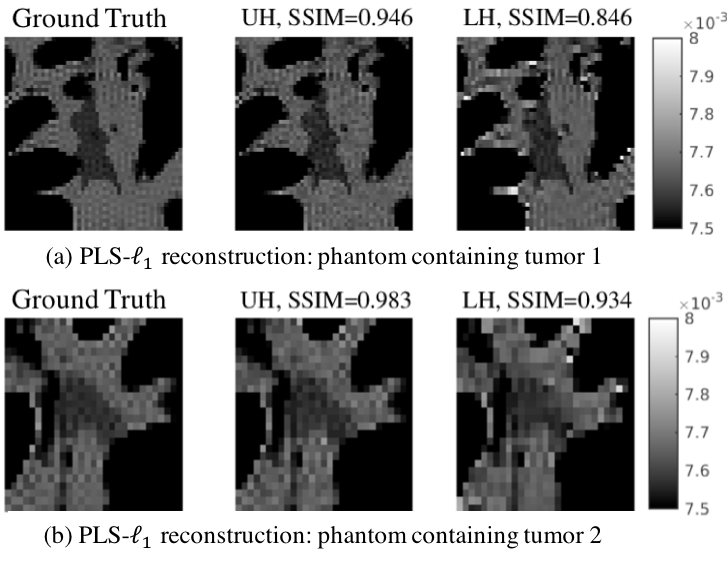}
 \caption{ROIs from images reconstructed by use of the PLS-$\ell_1$ estimator.
Subfigures (a) and (b) correspond to use of phantoms that contain tumors 1 and 2.
In each subfigure, the image in the left panel represents the phantom (ground truth).
The images in the center and right panels were reconstructed from data corresponding
to the UH and LH designs.}
 \label{fig:reconl1}
\end{figure}

To gain insights into the quality of the conflicting optimal half-scan data-acquisition designs identified by use
of the {SDO} and HO, images were reconstructed. 
Firstly, for each of the phantoms,
 PLS-$\ell_1$ image estimates were computed from the k-space data corresponding to
the {UH}  and {LH} data-acquisition designs. As described above,
these correspond to the optimal half-scan designs identified
by use of the {SDO} and HO, respectively.   The penalty term employed
in the  PLS-$\ell_1$ estimator was defined to be consistent with
 the sparse object prior in Eq.\ (\ref{eqn:prior}),
 as described in Sec.~\ref{sec:recon-demo}, {where the sparsifying transform was the Haar Wavelet transform introduced in Sec. \ref{sec:sd} for SDO test statistic calculations}. {The object was constrained to be real-valued.}  
 The reconstructed images are displayed
 in Fig.~\ref{fig:reconl1} and reveal that, in the case of the UH design, the
reconstructed tumor signal possesses have high contrast and clear boundaries.
On the other hand, in the case of the LH design, the
reconstructed tumor signal possesses lower contrast and distorted boundaries.
These observations suggest that the UH design that was identified as optimal by the {SDO}
may be a better choice for the specified signal{-}detection task than the LH design identified by the HO.
 The structural similarity (SSIM){\cite{wang2004image}} values, also displayed in Fig.~\ref{fig:reconl1},
 indicate that the images corresponding to the UH design are more similar to the true phantom
than those corresponding to the LH design.





\begin{figure*}[!t]
\centering
\includegraphics[width=\linewidth]{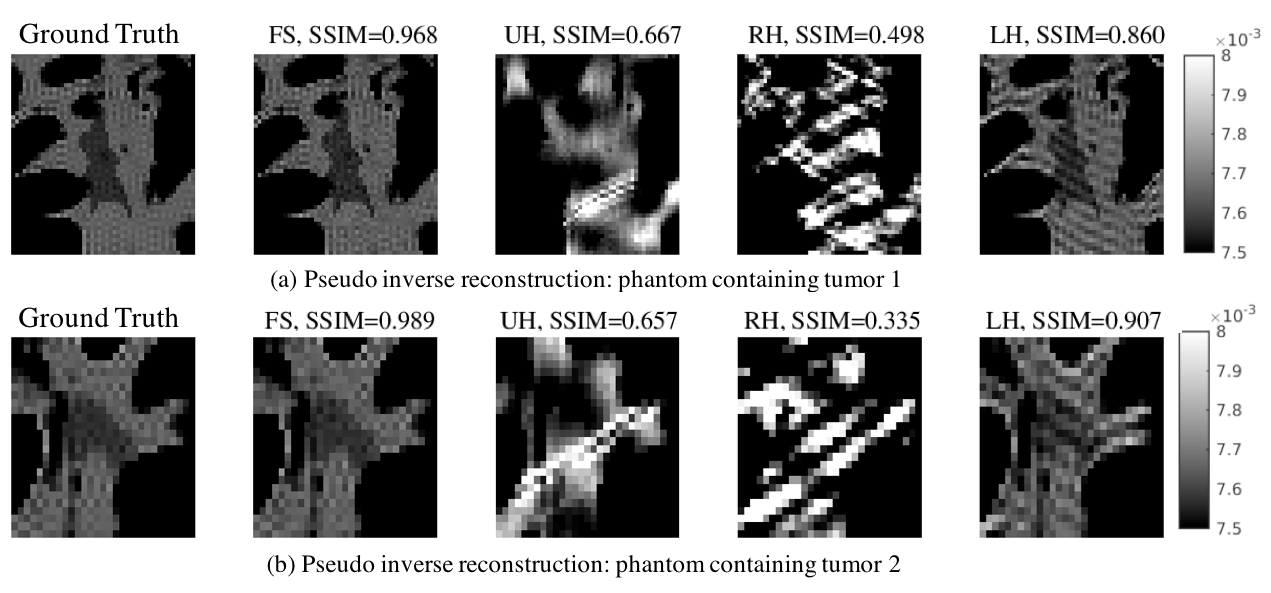}
 \caption{
ROIs from images reconstructed by use of the pseudoinverse reconstruction method.
Subfigures (a) and (b) correspond to use of phantoms that contain tumors  1 and 2.
In each subfigure, the image in the far-left panel represents the phantom (ground truth).
The remaining images, from left-to-right, were reconstructed from k-space data corresponding
to the FS, UH, RH, and LH, designs, respectively.}
 \label{fig:reconMP}
\end{figure*}




To further investigate the role of image reconstruction algorithm on design rank orderings,
images were re-reconstructed by use of a pseudoinverse method \cite{lustig2007sparse} instead of the sparse reconstruction method.
The reconstructed images  are displayed in Fig.~\ref{fig:reconMP}.
As observed with the PLS-$\ell_1$ estimator and reflected by the SSIM values,
 the images reconstructed from data corresponding to the FS design were the most similar to the true phantoms. 
{However, the visual appearance rankings of the images corresponding to the half-scan designs are distinct
from the the case where the PLS-$\ell_1$ estimator was employed.}
 Namely, in terms of visual appearances and SSIM values, the images corresponding to the LH design are now superior to the other designs.
Note that the images corresponding
 to the UH design (the optimal design according to the {SDO}), are highly distorted and therefore
likely to lead to compromised signal{-}detection performance.
This occurs because of a mismatch in the information that is utilized by the {SDO} and
pseudoinverse reconstruction method;  the {SDO} employed information regarding the
object sparsity that was not exploited by the pseudoinverse reconstruction method.
It should also be noted that the images reconstructed by use of the PLS-$\ell_1$ method 
corresponding to the UH design (Fig.~\ref{fig:reconl1}a and b, center panels) are more similar to the
phantoms than are the images reconstructed by use of the pseudoinverse
method (Fig.~\ref{fig:reconMP}a and b, right-most panel) corresponding to the LH design.
These observations support our conjecture that, when attempting to mitigate data incompleteness by use of a sparse image
reconstruction method, {matching statistical information
utilized by the observer and reconstruction method can be advantageous for optimizing system designs}.

\section{Discussion and conclusion}
\label{sec:discussion}


The {SDO} represents a new type of numerical observer that will find application
in optimizing the performance of modern computed imaging systems that employ
sparse reconstruction methods.  
Unlike a traditional IO that is predicated upon comprehensive statistical knowledge
of the class of objects to-be-imaged, the {SDO}
 assumes a stochastic {and data-driven} object model that describes only the sparsity properties of the class of objects.
{As such, the SDO and sparse reconstruction method
both utilize the same statistical information related to the sparsity
properties of the class of objects to-be-imaged.}
This information is already known to be useful for sparse reconstruction methods.
  As such, the use of the {SDO} will permit task-specific optimization of  data-acquisition
and system designs with consideration of the sparse reconstruction method to be employed.

An important contribution of this work was the development and implementation of
a method for computing the {SDO} test statistic for the case of i.i.d. Gaussian measurement noise. 
To circumvent the computational burden of MCMC methods, a variational method for approximating
the required posterior distribution was adopted.  Subsequently, the
test statistic could be computed semi-analytically{; this
can reduce  computation times by orders of magnitude compared with MCMC methods.}
 To the best of our knowledge, this represents the first application of a variational Bayesian inference method for estimating an {likelihood ratio} test statistic.  


{Many existing methods for imaging system optimization prescribe that imaging hardware or data-acquisition designs
 should be optimized by use of an IO that exploits full statistical knowledge of the measurement
noise and class of objects to-be-imaged, without consideration of the reconstruction method
 \cite{barrett2013foundations,zeng2012outcome}. }
The numerical studies presented in this work suggest that the notion may not always be applicable to modern imaging systems that employ sparse image
reconstruction methods.   Firstly, in a stylized MRI study, it was demonstrated that the {SDO} and HO
yielded different rank orderings of data-acquisition designs.   When a sparse 
reconstruction method was employed, the optimal design predicted by the {SDO} 
resulted in reconstructed images with decreased artifact levels as compared to
the images corresponding to the optimal HO-specified design.
 {Moreover, it is possible that  the optimal
 system design can change when the sparse reconstruction method is changed.}
Secondly, from the rank ordered data-acquisition designs described above, images
were re-reconstructed by use of a pseudo inverse method instead of the sparse reconstruction method.
In this case, the optimal design predicted by the HO
resulted in reconstructed images with significantly decreased artifact levels as compared to
the images corresponding to the optimal {SDO}-specified design.
These studies suggest that it can be important
to consider the image reconstruction method to be employed when
adopting a numerical observer to optimize the performance of computed
imaging systems that employ sparse reconstruction methods.
More specifically, in such applications, it can be important for the numerical
observer and reconstruction method to be statistically matched.

There remain numerous topics for future investigation.
In terms of technical developments, there is a need to extend the presented work to address other
measurement noise models { such as a generalized Gaussian
 noise model or a Poisson noise model. To accomplish this,
 additional approximations may be required and the impact of the approximations
on the accuracy of estimated posterior distribution should be studied systematically.}
 It may also be important to generalize the method to address
more complicated signal{-}detection tasks, such as signal-known-statistically tasks.
To further reduce the computational burden of the method for use with three-dimensional
imaging systems, a channelized version \cite{he2013model} of the {SDO} can also be explored.
Finally, it will be important to identify real-world applications for which imaging system performance
can be improved by use of the {SDO}.

\section*{Acknowledgments}

This research was supported in part by awards
NIH EB020168 and NIH EB020604 and NSF   DMS1614305.
The authors thank Drs.\ Craig Abbey, Jovan Brankov, Emil Sidky, and Frank Brooks 
for constructive comments that improved the clarity of our presentation.

\appendices
\section{Computation of the {SDO} Test Statistic}
\label{app:A}

The {SDO} test statistic was computed as described
by the pseudocode presented in Alg.~\ref{alg:SDIO}. First, the parameter $\tau$ for the Laplacian distribution
 was estimated based on the statistics of the training samples. Second, for each test sample, $\gamma_i$ ($i=1,2,...,Q$) values were chosen to minimize the KL-divergence between the approximated Gaussian distribution in Eq.\ (\ref{eqn:a_posterior}) and the true posterior distribution in Eq.\ (\ref{eqn:posterior}). The optimization problem can be solved by a double loop algorithm \cite{seeger2010optimization} shown in Alg.~\ref{alg:DL}. {The iterations should stop when the $\gamma_i$ values converge. In our implementation, a fixed maximum iteration number $k_0$ was empirically selected to be 16.} Finally, the likelihood ratios were calculated. 

{Note that a real-valued object model was assumed in this study. In other words, $\mathbf f\in\mathbb R^N$. Thus in the double-loop algorithm, the variance vector $\mathbf z$, object vector $\mathbf f$ and the matrix $\boldsymbol\Gamma$ corresponding the approximated prior are all real-valued. Extension to complex-valued object models is also possible by modeling the real and imaginary components of the objects separately by different Laplacian priors. More details are available in literature\cite{seeger2009speeding,seeger2010optimization}.}


\begin{algorithm}
\caption{{SDO} test statistic calculation for one test sample}\label{alg:SDIO}
\begin{algorithmic}[1]
\STATE Estimate the parameter $\tau$ for the Laplacian distribution based on the test samples
\STATE Calculate the $\gamma_i^{(0)}$ ($i=1,2,...,Q$) values by use of the double loop algorithm
\STATE Solve the optimization problem in Eq.\ (\ref{eqn:LR3-2})
\STATE Calculate the likelihood ratio based on Eq.\ (\ref{eqn:LR3})
\end{algorithmic}
\end{algorithm}

\begin{algorithm}
\caption{Double loop algorithm}\label{alg:DL}
\begin{algorithmic}[1]
\STATE Initialize $\gamma_i^{(0)}$ ($i=1,2,...,Q$) {to a large number ($1000$ in our implementation)} and form the diagonal matrix $\boldsymbol{\Gamma}^{(0)}$
\FOR{$k$=1,2,...{$k_0$}}
\STATE Calculate {$\mathbf A^{(k)}=\mathbf B^\dagger(\boldsymbol\Gamma^{(k-1)})^{-1}\mathbf B+\mathbf H^\dagger\boldsymbol\Sigma_n^{-1}\mathbf H$}
\STATE Calculate {$\mathbf z^{(k)}= $\,diag$^{-1}(\mathbf B[\mathbf A^{(k)}]^{-1}\mathbf B^\dagger)$}
\STATE Solve the optimization problem using a gradient descent algorithm (the inner-loop)
{\[\mathbf f^{(k)} = \arg\min_{\mathbf f}\sigma^{-2}\|\mathbf {Hf- g}\|^2+2\tau\sqrt{z_i^{(k)}+[\mathbf{Bf}]_i^2}\]}
\STATE Calculate {$[\gamma^{(k)}]_i = \tau^{-1}\sqrt{z_i^{(k)}+[\mathbf{B f}^{(k)}]_i^2}$}. Form the diagonal matrix $\boldsymbol \Gamma^{(k)}$
\ENDFOR
\end{algorithmic}
\end{algorithm}

%

\section{{Estimation of Laplacian Distribution Parameter}}
\label{app:tau}

{The Laplacian distribution parameter $\tau$ was chosen based on a training dataset. For the sparse prior assumed in Eq.\ (\ref{eqn:prior}), the variance of the Laplacian distribution is $2/\tau^2$. Thus, the Laplacian parameter was specified as
\begin{equation}
\label{eqn:wi2}
\tau = \sqrt{2/\text{var}(w)},
\end{equation}
where $w$ denotes an element in the transformed sparse vector $\mathbf w$ obtained by Eq.\ (\ref{eqn:sparsetransform}), and var$(w)$ denote the sample variance of $w$, estimated by 
\begin{equation}
\label{eqn:wi}
\text{var}(w) = E(w^2)-E(w)^2.
\end{equation}
Here, $E(w)$ and $E(w^2)$ denote the expectations of $w$ and $w^2$, respectively.
 Both of the expectations were empirically estimated based on a pool of sparse vector element samples formed by stacking together all the transformed vector elements from training images.}

{In our study, the wavelet transform was employed as the sparsifying transform.
 Thus, $w$ denotes a wavelet coefficient.
 Note that in this case,} there can be outliers that lead to underestimation of $\tau$.
 To circumvent this, the wavelet coefficients were subjected to
a threshold that removed outliers prior to estimating $\tau$ via Eq.\ (\ref{eqn:wi2}).
 {The threshold value was selected empirically. 
We varied $\tau$ within a range of values that yielded visually reasonable reconstructed images (data not shown); the conclusions related to the ranking of the data-acquisition designs was not altered.}

{\section{Inversion of a Large Matrix}}

The inversion of a large matrix can be a bottleneck in the computation
of numerical observers \cite{barrett2001megalopinakophobia}.
In the present case, the inversion of  $\mathbf A$ is the most
 computationally challenging task in the double loop algorithm. {The sparsifying matrix $\mathbf B$ is full rank and $\mathbf B^{-1} = \mathbf B^\dagger$. Thus we have $\mathbf B[\mathbf A^{(k)}]^{-1}\mathbf B^\dagger = [\mathbf B\mathbf A^{(k)}\mathbf B^\dagger]^{-1}$. {Denote $\mathbf C=\mathbf B\mathbf A^{(k)}\mathbf B^\dagger$}. It follows that
 $\mathbf C=[(\boldsymbol\Gamma^{(k-1)})^{-1}+\mathbf B\mathbf H^\dagger\boldsymbol\Sigma_n^{-1}\mathbf H\mathbf B^\dagger]^{-1}$. Thus,  steps 3 and 4 in Alg.~\ref{alg:DL} can be combined into one step, where the matrix $\mathbf C$ is inverted, and only the diagonal elements of the inverse matrix are needed. In this study, a parallel sparse direct solver called MUMPS \cite{MUMPS:1,MUMPS:2} was employed for this calculation. For the MRI application in our study, the matrix $\mathbf C$ contains many small values;
as such,  the $\mathbf C$ matrix can be thresholded to form
 a new sparse
 matrix $\mathbf C_{\text{approx}}$ that closely approximates the original $\mathbf C$ matrix.
Consequently,  the inverse of the approximated matrix $(\mathbf C_{\text{approx}})^{-1}$ can
  be regarded as a close approximation of the original matrix's inverse $\mathbf C^{-1}$.
 However, $\mathbf C_{\text{approx}}$ is much easier to invert because it is sparse. 
The MUMPS solver is designed purposefully to compute any desired element in the inverse of a sparse matrix and thus was employed to obtain the diagonal elements of $(\mathbf C_{\text{approx}})^{-1}$. 
In our study, a threshold of $0.01$ was employed.
 In practice, the diagonal elements of $\mathbf C$ are preserved in $\mathbf C_{\text{approx}}$ to prevent solving an ill-conditioned inverse problem.}

\section{Derivation of {SDO} test statistics (Eq.\ (18))}
\label{app:D}
Below, the derivation of the {SDO} test statistic considered in this work is provided.
As mentioned in Sec.~III-B, by approximating the Laplacian priors using parameterized Gaussian priors, the corresponding approximated posterior distribution is given by Eq.\ (15): 
{
\begin{align}
	\centering
	&P(\mathbf{f}_b | \mathbf{g}, \mathcal{H}_0)_{Q(\mathbf{f}_b;\boldsymbol{\gamma})} =Z_a^{-1}\mathcal N(\mathbf g|\mathbf H\mathbf f_b,\sigma^2\mathbf I)\exp\left(-{\mathbf w^T \boldsymbol{\Gamma}^{-1}\mathbf w}/2\right) 		\nonumber \\
	&= Z_a^{-1} (2 \pi)^{\frac{n}{2}} |\boldsymbol{\Sigma}_n|^{\frac{1}{2}} \exp \left\{ -\frac{1}{2} (\mathbf{g} - \mathbf{Hf}_b)^\dagger  \boldsymbol{\Sigma}_n^{-1}(\mathbf{g} - \mathbf{Hf}_b) \right\}		 \times \exp \left( \frac{1}{2} \mathbf{f}_b^\dagger \mathbf{B}^\dagger \boldsymbol{\hat{\Gamma}}(\mathbf{g})^{-1} \mathbf{B} \mathbf{f}_b \right) 					\nonumber \\
	&\propto \exp \left\{ -\frac{1}{2} (\mathbf{g} - \mathbf{Hf}_b)^\dagger \boldsymbol{\Sigma}_n^{-1} (\mathbf{g} - \mathbf{Hf}_b) - \frac{1}{2} \mathbf{f}_b^\dagger \mathbf{B}^\dagger \boldsymbol{\hat{\Gamma}}(\mathbf{g})^{-1} \mathbf{B f}_b \right\}     \nonumber \\
	&\propto \exp \left\{ -\frac{1}{2} \left[ \mathbf{g} \boldsymbol{\Sigma}_n^{-1} \mathbf{g} - 2 \mathbf{g}^{\dagger} \boldsymbol{\Sigma}_n^{-1} \mathbf{H f}_b + \mathbf{f}_b^\dagger \mathbf{H}^\dagger \boldsymbol{\Sigma}_n^{-1} \mathbf{H f}_b + \mathbf{f}_b^\dagger \mathbf{B}^\dagger \boldsymbol{\hat{\Gamma}}(\mathbf{g})^{-1} \mathbf{B f}_b \right] \right\}		\nonumber \\ 
	&\propto \exp \left( -\frac{1}{2} \mathbf{g}^\dagger \boldsymbol{\Sigma}_n^{-1} \mathbf{g} \right) 
			\exp \left\{ -\frac{1}{2} \left[ \mathbf{f}_b^\dagger \mathbf{H}^\dagger \boldsymbol{\Sigma}_n^{-1} \mathbf{H} \mathbf{f}_b 
			- 2 \mathbf{g}^\dagger \boldsymbol{\Sigma}_n^{-1} \mathbf{H f}_b + \mathbf{f}_b^\dagger \mathbf{B}^\dagger \boldsymbol{\hat{\Gamma}}(\mathbf{g})^{-1} \mathbf{B f}_b \right] \right\} 		\nonumber \\
	&\propto \exp \left\{ -\frac{1}{2} \left[ \mathbf{f}_b^\dagger \left( \mathbf{H}^\dagger \boldsymbol{\Sigma}_n^{-1} \mathbf{H} + \mathbf{B}^\dagger \boldsymbol{\hat{\Gamma}}(\mathbf{g})^{-1} \mathbf{B} \right) \mathbf{f}_b - 2 \mathbf{g}^\dagger \boldsymbol{\Sigma}_n^{-1} \mathbf{H f}_b \right] \right\}		\nonumber \\ 
	&\propto \exp \left[ -\frac{1}{2} (\mathbf{f}_b - \mathbf{f}_x)^\dagger \boldsymbol{\Sigma}(\mathbf{g})^{-1} (\mathbf{f}_b - \mathbf{f}_x) \right], \label{eq:app_posterior_exp}
\end{align}

where
\begin{equation}
	\left\{ \begin{array}{l}
			\boldsymbol{\Sigma}(\mathbf{g}) = \left[ \mathbf{H}^\dagger \boldsymbol{\Sigma}_n^{-1} \mathbf{H} + \mathbf{B}^\dagger \boldsymbol{\hat{\Gamma}}(\mathbf{g})^{-1} \mathbf{B} \right]^{-1}		\\
			\mathbf{f}_x = \boldsymbol{\Sigma}(\mathbf{g})^\dagger \mathbf{H}^\dagger \boldsymbol{\Sigma}_n^{-1} \mathbf{g}				\end{array}
	\right. 
\end{equation}
and terms independent of $\mathbf f_b$ are omitted in the derivation. 
Based on the fact that the distribution is normalized, 
\begin{equation}
P(\mathbf{f}_b | \mathbf{g}, \mathcal{H}_0)_{Q(\mathbf{f}_b;\boldsymbol{\gamma})} = Z_1^{-1}\exp \left[ -\frac{1}{2} (\mathbf{f}_b - \mathbf{f}_x)^\dagger \boldsymbol{\Sigma}(\mathbf{g})^{-1} (\mathbf{f}_b - \mathbf{f}_x) \right], \label{eq:app_posterior_exp_final}
\end{equation}
where $Z_1 = (2 \pi)^{\frac{n}{2}} \left\vert \boldsymbol{\Sigma}(\mathbf{g}) \right\vert ^{\frac{1}{2}}$.}

The BKE likelihood ratio is given by Eq.\ (11): 
\begin{equation}
	\Lambda_{BKE}(\mathbf{g} \vert \mathbf{f}_b) = \exp \left[ \left( \mathbf{g} - \mathbf{H f}_b - \frac{1}{2} \mathbf{H f}_s \right)^\dagger \boldsymbol{\Sigma}_n^{-1} \mathbf{H f}_s \right].		
	\label{eq:app_BKE}
\end{equation}
{By taking Eq.\ (\ref{eq:app_posterior_exp_final}) and Eq.\ (\ref{eq:app_BKE}) into the integral form of ideal observer test statistics in Eq.\ (4), one obtains
\begin{align}
	\centering
	&\Lambda(\mathbf{g}) = \int \diff\mathbf{f}_b \Lambda_{BKE}(\mathbf{g} \vert \mathbf{f}_b) P(\mathbf{f}_b \vert \mathbf{g}, \mathcal{H}_0)_{Q(\mathbf{f}_b; \boldsymbol{\gamma})}		\nonumber \\
&= \int \diff\mathbf{f}_b \exp\left\{ \left( \mathbf{g} - \mathbf{H f}_b - \frac{1}{2} \mathbf{H f}_s \right)^\dagger \boldsymbol{\Sigma}_n^{-1} \mathbf{H f}_s \right\} Z_1^{-1} \exp \left\{ -\frac{1}{2} (\mathbf{f}_b - \mathbf{f}_x)^\dagger \boldsymbol{\Sigma}(\mathbf{g})^{-1} (\mathbf{f}_b - \mathbf{f}_x) \right\}		\nonumber \\
			 &= Z_1^{-1} \int \diff\mathbf{f}_b \exp \left\{ \left( \mathbf{g} - \frac{1}{2}\mathbf{H f}_s \right)^\dagger \boldsymbol{\Sigma}_n^{-1} \mathbf{H f}_s - (\mathbf{H f}_b)^\dagger \boldsymbol{\Sigma}_n^{-1} \mathbf{H f}_s \right\} 		\nonumber \\
		 &\qquad   \exp \left\{ -\frac{1}{2} \left[ \mathbf{f}_b^\dagger \boldsymbol{\Sigma}(\mathbf{g})^{-1} \mathbf{f}_b - \mathbf{f}_b^\dagger \boldsymbol{\Sigma}(\mathbf{g})^{-1} \mathbf{f}_x - \mathbf{f}_x^\dagger \boldsymbol{\Sigma}(\mathbf{g})^{-1} \mathbf{f}_b + \mathbf{f}_x^\dagger \boldsymbol{\Sigma}(\mathbf{g})^{-1} \mathbf{f}_x  \right]  \right\} \nonumber \\
		&= Z_1^{-1} \exp \left[ (\mathbf{g} - \frac{1}{2} \mathbf{H f}_s)^\dagger \boldsymbol{\Sigma}_n^{-1} \mathbf{H f}_s \right] \exp \left[ -\frac{1}{2} \mathbf{f}_x^\dagger \boldsymbol{\Sigma}(\mathbf{g})^{-1} \mathbf{f}_x \right] 	\nonumber\\
		&\qquad \times\int \diff\mathbf{f}_b \exp \left\{ -\frac{1}{2} \left[ \mathbf{f}_b^\dagger \boldsymbol{\Sigma}(\mathbf{g})^{-1}\mathbf{f}_b  
	- 2 \mathbf{f}_x^\dagger \boldsymbol{\Sigma}(\mathbf{g})^{-1} \mathbf{f}_b+ 2(\mathbf{H f}_s)^\dagger \boldsymbol{\Sigma}_n^{-1} \mathbf{H f}_b  \right]  \right\}. 		\label{eq:app_LR_temp_1}
\end{align}
Let $A = \exp \left[ (\mathbf{g} - \frac{1}{2} \mathbf{H f}_s)^\dagger \boldsymbol{\Sigma}_n^{-1} \mathbf{H f}_s \right] \exp \left[ -\frac{1}{2} \mathbf{f}_x^\dagger \boldsymbol{\Sigma}(\mathbf{g})^{-1} \mathbf{f}_x \right]$:
\begin{align}
	\centering
	\Lambda(\mathbf{g}) &= Z_1^{-1} A \int \diff\mathbf{f}_b \exp \left\{ -\frac{1}{2} \left[ \mathbf{f}_b^\dagger \boldsymbol{\Sigma}(\mathbf{g})^{-1} \mathbf{f}_b - 2 \left( \mathbf{f}_x^\dagger \boldsymbol{\Sigma}(\mathbf{g})^{-1} - (\mathbf{H f}_s)^\dagger \boldsymbol{\Sigma}_n^{-1} \mathbf{H} \right) \mathbf{f}_b \right]  \right\} \nonumber \\
			&= Z_1^{-1} A \int \diff\mathbf{f}_b \exp \left\{ -\frac{1}{2} \left[ \mathbf{f}_b^\dagger \boldsymbol{\Sigma}(\mathbf{g})^{-1} \mathbf{f}_b - 2 \mathbf{f}_y^\dagger \boldsymbol{\Sigma}(\mathbf{g})^{-1} \mathbf{f}_b \right] \right\}		\label{eq:app_LR_temp_2}
\end{align}
where $\mathbf{f}_y = \mathbf{f}_x - \boldsymbol{\Sigma}(\mathbf{g})^\dagger \mathbf{H}^\dagger \boldsymbol{\Sigma}_n^{-1} \mathbf{H f}_s$. 
Equation (\ref{eq:app_LR_temp_2}) can further be written as: 
\begin{align}
	\Lambda(\mathbf{g}) &= Z_1^{-1} A \int \diff \mathbf{f}_b \exp \left\{ -\frac{1}{2} \left[ \mathbf{f}_b^\dagger \boldsymbol{\Sigma}(\mathbf{g})^{-1} \mathbf{f}_b - 2 \mathbf{f}_y^\dagger \boldsymbol{\Sigma}(\mathbf{g})^{-1} \mathbf{f}_b + \mathbf{f}_y^\dagger \boldsymbol{\Sigma}(\mathbf{g})^{-1} \mathbf{f}_y  \right] \right\} 
	\exp \left[ \frac{1}{2} \mathbf{f}_y^\dagger \boldsymbol{\Sigma}(\mathbf{g})^{-1} \mathbf{f}_y \right] 		\nonumber \\
	&= Z_1^{-1} A \exp \left[ \frac{1}{2} \mathbf{f}_y^\dagger \boldsymbol{\Sigma}(\mathbf{g})^{-1} \mathbf{f}_y \right]   \times\int \diff\mathbf{f}_b \exp \left[ -\frac{1}{2} (\mathbf{f}_b -\mathbf{f}_y)^\dagger \boldsymbol{\Sigma}(\mathbf{g})^{-1} (\mathbf{f}_b - \mathbf{f}_y) \right]		\nonumber \\
	&=  A \exp \left[ \frac{1}{2} \mathbf{f}_y^\dagger \boldsymbol{\Sigma}(\mathbf{g})^{-1} \mathbf{f}_y \right],	\label{eq:app_LR_temp_3}
\end{align}}
Therefore, Eq.\ (\ref{eq:app_LR_temp_3}) can be further simplified as: 
\begin{align}
\Lambda(\mathbf{g}) 
			= \exp \Biggl\{& (\mathbf{g} - \frac{1}{2} \mathbf{H f}_s)^\dagger \boldsymbol{\Sigma}_n^{-1} \mathbf{H f}_s+ \frac{1}{2} \big[ (\mathbf{g} - \mathbf{H f}_s)^\dagger \boldsymbol{\Sigma}_n^{-1} \mathbf{H} \boldsymbol{\Sigma}(\mathbf{g}) \mathbf{H}^\dagger \boldsymbol{\Sigma}_n^{-1} (\mathbf{g} - \mathbf{H f}_s)		- \mathbf{g}^\dagger \boldsymbol{\Sigma}_n^{-1} \mathbf{H} \boldsymbol{\Sigma}(\mathbf{g}) \mathbf{H}^\dagger \boldsymbol{\Sigma}_n^{-1} \mathbf{g} \big] \Biggr\} \nonumber \\
					= \exp \biggl\{ &(\mathbf{g} - \frac{1}{2} \mathbf{H f}_s)^\dagger [\mathbf{I} - \boldsymbol{\Sigma}_n^{-1} \mathbf{H} \boldsymbol{\Sigma}(\mathbf{g}) \mathbf{H}^\dagger] \boldsymbol{\Sigma}_n^{-1} \mathbf{H f}_s   \biggr\},		
\end{align}
which is same as Eq.\ (18).


\ifCLASSOPTIONcaptionsoff
  \newpage
\fi



\bibliographystyle{IEEEtran}
\bibliography{SDO_arxiv}
\end{document}